\documentclass{article}

    \PassOptionsToPackage{numbers, compress}{natbib}

\usepackage[main, final]{neurips_2026}
\usepackage{enumitem}

\usepackage[utf8]{inputenc} 
\usepackage[T1]{fontenc}    
\usepackage{hyperref}       
\usepackage{url}            
\usepackage{multirow}
\usepackage{booktabs}       
\usepackage{makecell}
\usepackage{wrapfig}
\usepackage{amsfonts}       
\usepackage{amsmath}
\usepackage{amssymb}
\usepackage{nicefrac}       
\usepackage{microtype}      
\usepackage[table]{xcolor}         
\usepackage[most]{tcolorbox}
\usepackage{graphicx}
\usepackage{fontawesome5}

\usepackage{algorithm}
\usepackage{algorithmic}

\newtcolorbox{mybox}[2][]{colbacktitle=red!10!white, colback=blue!10!white,coltitle=red!70!black, title={#2},fonttitle=\bfseries,#1}

\newcommand{\E}{\mathbb{E}}
\newcommand{\Var}{\mathrm{Var}}

\def\sD{{\mathbb{D}}}
\newcommand{\train}{\mathcal{D}}
\newcommand{\Ls}{\mathcal{L}}
\newcommand{\Lpolicy}{\mathcal{L}_{\rm pol}}
\DeclareMathOperator*{\argmin}{arg\,min}

\title{Staying in the Sweet Spot: Boost Reasoning Evolution via Responsive Rollout Difficulty Manipulation}

\author{Ziheng Li$^{1}$\thanks{The first two authors contribute equally.}\hspace{0.4em}\thanks{Work done during internship at Baidu.}, Zexu Sun$^{2}$\footnotemark[1], Jinman Zhao$^{3}$, Erxue Min$^{2}$, Yongcheng Zeng$^{4}$, Hui Wu$^{5}$, \\ \textbf{Hengyi Cai$^{2}$, Shuaiqiang Wang$^{2}$, Dawei Yin$^{2}$, Xu Chen$^{6}$\footnotemark[3], Zhi-Hong Deng$^{1}$\thanks{Corresponding authors.}}
\\
$^1$ State Key Laboratory of General Artificial Intelligence,\\
School of Intelligence Science and Technology, Peking University\\
$^2$ Baidu Inc. 
$^3$ Department of Computer Science, University of Toronto\\
$^4$ Institute of Automation, Chinese Academy of Sciences \\
$^5$ Aerospace Information Research Institute, Chinese Academy of Sciences\\
$^6$ Gaoling School of Artificial Intelligence, Renmin University of China \\
\texttt{\{liziheng, zhdeng\}@pku.edu.cn, sunzexu0826@gmail.com,}\\ \texttt{xu.chen@ruc.edu.cn}}

\makeatletter
\renewcommand{\@notice}{}
\makeatother

\begin{document}

\maketitle

\begin{abstract}
{\parfillskip=0pt
Reinforcement learning with verifiable rewards (RLVR) has advanced the reasoning capabilities of large language models (LLMs).
However, existing RLVR methods often suffer from exploration inefficiency due to mismatches between problem difficulty and model capability: overly difficult problems hinder reasoning path discovery, while overly simple problems offer little learning signal.
To address this, we first formalize the effect of problem difficulty by quantifying the relationship between loss descent magnitude and rollout accuracy.
Building on this analysis, we propose \textbf{SEELE}, a supervision-aided RLVR framework that dynamically adjusts problem difficulty to lie within the high-performance region.
SEELE augments each training sample by appending a hint (part of a full solution) for difficulty reduction. 
Unlike previous hint-based approaches, SEELE deliberately computes the hint length for each individual problem to achieve an optimal difficulty.
The optimal hint length is determined via multi-round rollout sampling, where an item response theory model fits accuracy–hint pairs from previous rounds to predict the next-round hint.
This instance-level, real-time difficulty adjustment aligns problem difficulty with the evolving model capability, thereby improving exploration efficiency. 
Experiments show that SEELE outperforms Group Relative Policy Optimization (GRPO) and Supervised Fine-tuning (SFT) by \textbf{+10.0} and \textbf{+8.4} points, respectively, and exceeds the best prior supervision-aided approach by \textbf{+3.8} points on average across six math reasoning benchmarks.
\par}

\vspace{5pt}
\faGithub~\textbf{GitHub:} \href{https://github.com/ChillingDream/seele}{https://github.com/ChillingDream/seele}
\end{abstract}

\section{Introduction}
Recent large language models (LLMs) such as GPT-5~\citep{singh2025openai}, DeepSeek-R1~\citep{deepseek-aiDeepSeekR1IncentivizingReasoning2025}, Kimi-K2~\citep{teamKimiK2Open2025}, and GLM-5~\citep{zeng2026glm} have made remarkable breakthroughs in reasoning ability, benefiting from long chain-of-thought that incorporates self-reflection and revision.
Prior works~\citep{deepseek-aiDeepSeekR1IncentivizingReasoning2025} have shown that the capability for complex reasoning can be realized through pure reinforcement learning with verifiable rewards (RLVR), which forgoes memorizing annotated reasoning processes and instead elicits models' inherent capabilities by extensive on-policy exploration and strengthening verified correct trajectories~\citep{chuSFTMemorizesRL2025}.
At present, RLVR has become the common practice for building high-performance reasoning models.

However, on-policy exploration inherently constrains learning efficiency, exhibiting strong data dependency~\citep{gaoNavigateUnknownEnhancing2025,douImprovingRLExploration2025,schmiedLLMsAreGreedy2025,yuDAPOOpenSourceLLM2025,zhang2025survey,sun2025uncertainty}.
RLVR is driven by the rewards from extensive online sampled rollouts, which collapse to zero when the training problems are too difficult for LLMs to produce a correct response.
Conversely, overly simple problems yield nearly all correct rewards, producing minor advantage value.
It remains unclear what level of problem difficulty maximizes learning efficiency and how to curate such data.
Moreover, as recent studies~\citep{gandhiCognitiveBehaviorsThat2025,yueDoesReinforcementLearning2025,zhaoEchoChamberRL2025} have found, RLVR merely amplifies existing behaviors rather than fostering novel cognitive capabilities, thereby limiting the achievable performance to that of the base model. 
Supervised fine-tuning (SFT)~\citep{kopf2023openassistant} is a naive way to improve the ability of LLMs before RL with expert data. However, existing works~\citep{zhang2025policy,chen2025sft} show that directly using SFT-then-RL is not an effective way, which even underperforms pure RL. 

\begin{figure*}[t]
    \centering
    \includegraphics[width=0.98\linewidth]{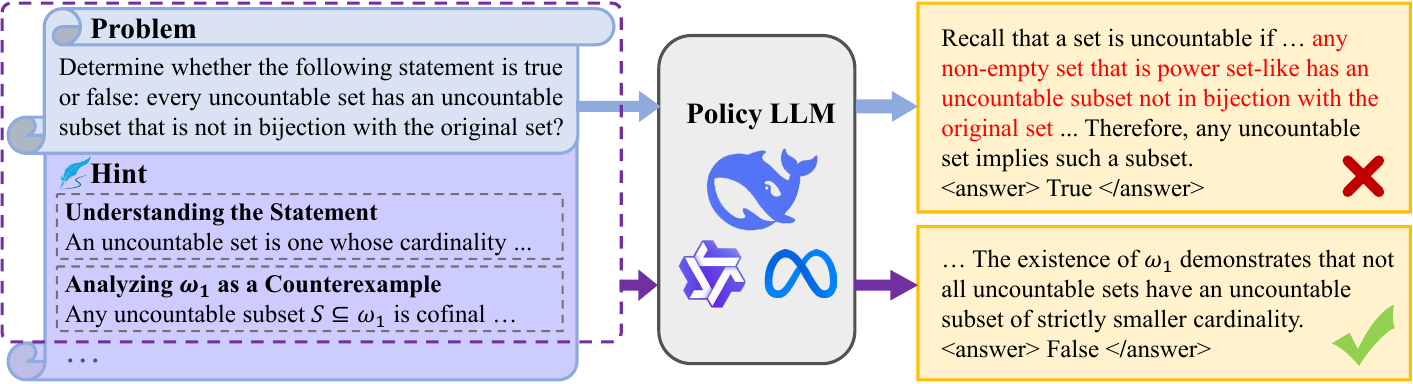}
    \caption{Comparison between the direct rollout (blue) and hinted rollout (purple). The hint consists of the first few steps from an annotated solution. Adding a hint can simplify the problem and guide the LLM toward completing correct solutions.}
    \label{fig:hint_intro}
\end{figure*}

To overcome these limitations, recent works have attempted to integrate SFT into the RL framework, enabling synergistic learning when SFT-then-RL is ineffective. 
LUFFY~\citep{yanLearningReasonOffPolicy2025} and SRFT~\citep{fuSRFTSingleStageMethod2025} incorporate parallel off-policy guidance into the rollout set, allowing simultaneous exploration and imitation. 
UFT~\citep{liuUFTUnifyingSupervised2025}, TAPO~\citep{wuThoughtAugmentedPolicyOptimization2025}, StepHint~\citep{zhangStepHintMultilevelStepwise2025}, Hint-GRPO~\citep{huangBoostingMLLMReasoning2025}, and Prefix-RFT~\citep{huangBlendingSupervisedReinforcement2025} append an off-policy hint prefix to each problem to reduce exploration difficulty. 
While these methods allow for a certain degree of problem difficulty modulation, they have a salient shortcoming: off-policy guidance is applied statically and indiscriminately across problems and hint levels, causing most training samples' difficulty to mismatch the model’s evolving capability.
Consequently, for those challenging problems, mild off-policy guidance does not effectively resolve the low efficiency of on-policy exploration, whereas for those easy problems, excessive off-policy intervention may impede the LLM from developing its own reasoning patterns.
This raises a critical research question: 
\begin{center}
\emph{What is the appropriate problem difficulty, and how can the difficulty be dynamically adjusted in accordance with the model's evolving capability?} 
\end{center}

In this paper, we propose \textbf{SEELE}: \textit{\textbf{S}weet-spot st\textbf{EE}ring \textbf{LE}arning}, a theoretically-grounded supervision-aided RLVR approach that keeps high learning efficiency throughout the whole training stage via dynamically adjusting the proportion of the off-policy prefixes.
SEELE explicitly formalizes that an appropriately difficult problem should yield a rollout accuracy of approximately 50\% through a theoretical analysis of the loss descent magnitude.
By appending an \emph{instance-specific, dynamically determined} hint after the original problem (Figure~\ref{fig:hint_intro}), SEELE is able to control the problem difficulty within the ``sweet spot''.
Unlike previous hint-based methods~\citep{liuUFTUnifyingSupervised2025,huangBlendingSupervisedReinforcement2025}, our approach enables per-instance adaptation and involves the model's real-time feedback as training progresses, thereby achieving a more precise alignment between problem difficulty and model capability.
Concretely, we split a rollout sampling stage into several rounds, across which we establish a regression model to predict the accuracy given the hinting rate (proportion of the full solution) under the guidance of item response theory~\citep{chenItemResponseTheory2021}.
At each round, we fit the accuracy prediction model using the feedback from the previous rounds and predict how long the current hint should be for a 50\% accuracy.
We conduct experiments on six math reasoning benchmarks and three out-of-domain general reasoning benchmarks, on which SEELE significantly outperforms previous RLVR methods.

Our contributions can be summarized as:
\begin{itemize}[leftmargin=*]
    \item We theoretically show that the upper bound on the learning efficiency of RLVR algorithms follows a quadratic relationship with rollout accuracy, peaking at 50\% accuracy.
    \item Guided by our theoretical analysis, we propose a novel capability-adaptive RLVR framework that manipulates the problem difficulty via multi-round rollout sampling and accuracy prediction, maintaining high learning efficiency throughout the entire training process.
    \item We demonstrate SEELE's superiority over previous RLVR methods on 9 challenging benchmarks, achieving remarkable improvements of \textbf{+10.0} points compared with GRPO on math reasoning.
\end{itemize}

\section{Related Work}
\textbf{Reinforcement Learning for LLM Reasoning.}
Recent work has demonstrated the effectiveness of reinforcement learning (RL) in improving the reasoning capabilities of large language models (LLMs), as exemplified by systems such as DeepSeek-R1~\citep{deepseek-aiDeepSeekR1IncentivizingReasoning2025}, OpenAI-o1~\citep{openaiOpenAIO1System2024}, and Kimi-K2~\citep{teamKimiK2Open2025}.
These studies show that pure RL with verifiable rewards can drive LLMs to autonomously develop extended chain-of-thought reasoning patterns, incorporating substantial self-reflection and iterative refinement.
Among these approaches, GRPO-based methods have emerged as a pivotal paradigm for enhancing reasoning performance.
Subsequent studies such as DAPO~\citep{yuDAPOOpenSourceLLM2025}, Dr.GRPO~\citep{liuUnderstandingR1ZeroLikeTraining2025}, GSPO~\citep{zhengGroupSequencePolicy2025} focus on addressing GRPO's optimization limitations by removing length bias, difficulty bias, etc.
However, recent works~\citep{yueDoesReinforcementLearning2025,wangReinforcementLearningReasoning2025} argue that RL approaches are bounded by the capabilities of the base model and do not acquire new reasoning skills, which suggests pure RL may not be the ultimate solution.

\textbf{Supervision-Aided Reinforcement Learning.}
Supervised fine-tuning (SFT) on high-quality reasoning chains effectively strengthens reasoning ability and is typically applied before the RL stage~\citep{openaiOpenAIO1System2024,qwenQwen25TechnicalReport2025}.  
To overcome efficiency and capacity limits of RL, later studies integrate off-policy supervision into RL as a single process.  
LUFFY~\citep{yanLearningReasonOffPolicy2025} enriches RL rollouts with off-policy annotated traces for external guidance, while SRFT~\citep{fuSRFTSingleStageMethod2025} jointly optimizes SFT and RL losses using an entropy-based weighting scheme.  
UFT~\citep{liuUFTUnifyingSupervised2025}, StepHint~\citep{zhangStepHintMultilevelStepwise2025}, and Hint-GRPO~\citep{huangBoostingMLLMReasoning2025} append partial solutions (hints) from stronger models to handle difficult problems.
Prefix-RFT~\citep{huangBlendingSupervisedReinforcement2025} similarly employs hints but excludes low-entropy hint tokens to avoid over-imitation.  
TAPO~\citep{wuThoughtAugmentedPolicyOptimization2025} instead introduces high-level ``thought patterns'' to promote external strategy learning.  
Yet these methods rely on static supervision, limiting exploration and imitation when tasks are mismatched to the model’s capability.  
In contrast, SEELE grounds hint assignment in principled RL optimization and introduces a capability-adaptive mechanism, enabling dynamic and explicit difficulty adaptation.

\textbf{Curriculum Learning.} Curriculum learning (CL) has been widely studied as a principled strategy for improving optimization by presenting training data in a meaningful order, typically from easy to hard, as first formalized by \citet{bengio2009curriculum}. Recent work extends this idea to reinforcement learning for LLM reasoning. \citet{parashar2025curriculum} apply a static easy-to-hard task schedule, while \citet{jiang2509vcrl,chen2025self} develop dynamic curricula based on reward variance or bandit-based data selection during RL fine-tuning. In parallel, \citet{zhang2025learning} continuously re-estimate problem difficulty to match the model’s evolving capabilities. However, existing approaches mainly rely on heuristic sample selection and therefore cannot achieve precise difficulty matching, while repeated resampling also sacrifices data efficiency. In contrast, our method provides a more direct and efficient solution for capability-aware difficulty control.

\section{Preliminaries}
\textbf{Reinforcement Learning with Verifiable Rewards (RLVR)} formulates the generation process of an LLM as a Markov Decision Process (MDP), where the state is defined as the concatenation of the prompt $x$ and the tokens generated so far $y_{1:t-1}$, and the action corresponds to selecting the next token $y_t$ from the policy, i.e., $y_t \sim \pi_\theta(\cdot|x,y_{1:t-1})$. 
The objective of RLVR is to train the policy model to generate outputs that achieve high scores under a rule-based reward function $r(x,y)$. 
Formally, RLVR optimizes the expected reward over the on-policy rollouts generated from the policy model $\pi_{\theta_{\rm old}}$ at the previous step, and the objective can be written as
\begin{equation}\label{eq:obj}
    \Ls(\theta)
    =\underbrace{-\E_{x\sim \train,\,y\sim\pi_\theta(\cdot|x)}\!\big[A_{\theta_{\rm old}}(x,y)\big]}_{\Lpolicy(\theta)}
    +\beta\underbrace{\E_{x\sim \train}\!\left[\sD_{\rm KL}\!\left(\pi_{\theta_{\rm ref}}(\cdot|x)\,\|\,\pi_\theta(\cdot|x)\right)\right]}_{\Ls_{\rm KL}(\theta)},
\end{equation}
where $A_{\theta_{\rm old}}(x,y)=r(x,y)-\E_{y\sim\pi_{\theta_{\rm old}}(\cdot|x)}[r(x,y)]$ denotes the advantage function, pol stands for ``policy'', and the hyperparameter $\beta$ controls the strength of KL regularization with respect to the reference model $\pi_{\rm ref}$.

\textbf{GRPO} offers an elegant advantage function implementation, which has been widely applied in RLVR studies~\citep{shaoDeepSeekMathPushingLimits2024}. Given an input $x$, GRPO samples a group of outputs $\{y_1,y_2,\cdots,y_n\}$ from $\pi_{\theta_{\rm old}}(\cdot|x)$ and uses the group mean to replace the expectation.
Recently, \citet{liuUnderstandingR1ZeroLikeTraining2025} propose an improved advantage function in their Dr.GRPO approach as
\begin{equation}
    A(x,y_i)=r(x,y_i)-{\rm mean}(\{r(x,y_j)|j=1,2,\cdots,n\}),
\end{equation}
where the normalization term is removed to mitigate optimization bias. 
In the following, our analysis is conducted based on this unnormalized formulation.

\section{Methodology}
In this section, we present SEELE from three aspects:
(1) a theoretical foundation that identifies the optimal problem difficulty in terms of rollout accuracy (\S~\ref{sec:theory});  
(2) a multi-round sampling framework that decomposes rollout generation into sequential rounds, thereby enabling capability-aware adaptation of problem difficulty (\S~\ref{sec:framework});
(3) a rollout accuracy prediction model based on Item Response Theory, which supports accurate and responsive adjustment of problem difficulty (\S~\ref{sec:irt}).

\subsection{Relationship Between Learning Efficiency and Rollout Accuracy}\label{sec:theory}
We begin by formulating a quantitative relationship between reinforcement learning (RL) training efficiency and problem difficulty.
The model prediction accuracy is employed as the difficulty measure.
Given a policy model $\pi_\theta$ and a binary reward function $r(x,y)$, for a problem $x$, we define the prediction accuracy for $x$ with respect to the policy model $\pi_\theta$ as follows:
\begin{equation}
    a_\theta(x)=\E_{y\sim\pi_\theta(\cdot|x)}[r(x,y)].
\end{equation}
Prediction accuracy is the expectation of the rollout accuracy and reflects the difficulty of this training instance relative to the current model capability.

Next, we consider a one-step gradient descent from the last-step policy model $\pi_{\theta_{\rm old}}$ with the update vector $d$.
For analytical convenience, we assume $\theta_{\rm ref} = \theta_{\rm old}$.
The minimization objective (Eq.~\eqref{eq:obj}) can be reformulated by letting $\theta=\theta_{\rm old}+d$:
\begin{equation}\label{eq:one-step}
    \min_d\Ls(\theta_{\rm old}+d)=\min_d\left\{\Lpolicy(\theta_{\rm old}+d)+\beta\E_{x\sim \train}\left[\sD_{\rm KL}\left(\pi_{\theta_{\rm old}}(\cdot|x)||\pi_{\theta_{\rm old}+d}(\cdot|x)\right)\right]\right\}.
\end{equation}

We conclude that the magnitude of loss descent is upper bounded by the quadratic envelope of $a_\theta(x)$:
\begin{equation}
    \Ls(\theta_{\rm old})-\Ls(\theta_{\rm old}+d)\leq\frac{1}{2\beta}\E_{x\sim\train}[a_{\theta_{\rm old}(x)}(1-a_{\theta_{\rm old}}(x))].\label{eq:upper-bound}
\end{equation}

Here, we briefly outline the derivation.
Since $\Ls$ is defined as the sum over independent prompts $x$, we analyze the instance-level loss descent $\Ls(x;\theta_{\rm old})-\Ls(x;\theta_{\rm old}+d)$ and then aggregate across all instances. 
From Eq.~\eqref{eq:one-step}, for a specific instance $x$, its minimizer $d_x^*$ can be approximated by applying first-order Taylor expansion on $\Lpolicy$ and second-order Taylor expansion on $\Ls_{\rm KL}$ at $\theta_{\rm old}$:
\begin{equation}\label{eq:approx}
    d_x^*\approx\argmin_d\left\{\Lpolicy(x;\theta_{\rm old})+\nabla_\theta\Lpolicy(x;\theta)\big|_{\theta=\theta_{\rm old}}^Td+\frac{\beta}{2}d^TF(\theta_{\rm old})d\right\},
\end{equation}
where $F(\theta_{\rm old})$ is the Fisher Information Matrix.
Since $F(\theta_{\rm old})$ is always positive semi-definite, the right side of Eq.~\eqref{eq:approx} is convex and has a global minimizer\footnote{The policy gradient lies in the range of $F$. If $F$ is singular, $F^{-1}$ denotes its inverse restricted to this range.}:
\begin{equation}
    d_x^*=-\frac{1}{\beta}F^{-1}(\theta_{\rm old})\nabla_\theta\Lpolicy(x;\theta_{\rm old}).
\end{equation}

By substituting $d_x^*$ into the Taylor expansion of $\Ls(x;\theta_{\rm old}+d)$, we derive the loss descent value:
\begin{align}
    \Ls(x;\theta_{\rm old})-\Ls(x;\theta_{\rm old}+d_x^*)
    \approx&\frac{1}{2\beta}\nabla_\theta\Lpolicy(x;\theta)\big|_{\theta=\theta_{\rm old}}^TF^{-1}(\theta_{\rm old})\nabla_\theta\Lpolicy(x;\theta)\big|_{\theta=\theta_{\rm old}}\\
    =&\frac{1}{2\beta}\nabla_\theta a_{\theta}(x)\big|_{\theta=\theta_{\rm old}}^TF^{-1}(\theta_{\rm old})\nabla_\theta a_{\theta}(x)\big|_{\theta=\theta_{\rm old}}.\label{eq:aFa}
\end{align}

Finally, note that $r(x,y)$ is an unbiased estimator of $a_\theta(x)$.
By applying the vector parameter Cramér–Rao bound to Eq.~\eqref{eq:aFa} and summing over $x\sim\train$, we get the upper bound shown in Eq.~\eqref{eq:upper-bound} (the equality becomes an inequality because $d$ may not simultaneously satisfy all $d_x^*$).
The full derivation and extension to the multi-step clipped PPO objective are provided in Appendix~\ref{sec:derivation}. 
Eq.~\eqref{eq:upper-bound} indicates the convergence rate is correlated with the problem difficulty.
The policy model learns little from too easy ($a_\theta(x)\to1$) or too hard problems ($a_\theta(x)\to0$), and the maximal efficiency upper bound is achieved at 50\% accuracy.

\subsection{Difficulty-Aware Hint Manipulation via Multi-Round Sampling}\label{sec:framework}
From Eq.~\eqref{eq:upper-bound}, we have established how problem difficulty affects learning efficiency. 
A natural follow-up problem is whether we can deliberately adjust the problem difficulty to lie within the high-efficiency region (around 50\%). 
Recent studies~\citep{liuUFTUnifyingSupervised2025,huangBlendingSupervisedReinforcement2025,yanLearningReasonOffPolicy2025,fuSRFTSingleStageMethod2025} have shown that incorporating off-policy hint guidance into on-policy exploration can enhance the success rate of exploration on challenging samples. 
However, their strategies for controlling hint length are purely heuristic and lack a principled objective, failing to consider the instance-level difficulty as well as the model’s real-time capability. 
Consequently, the difficulty of hinted problems may deviate from the optimal region, thereby limiting optimization efficiency.

We draw inspiration from these hint-based methods and propose SEELE, a novel instance-level capability-adaptive hint manipulation approach built on GRPO.
SEELE adds a dynamic length hint after the original problem to control difficulty to match the increasing model capability, maintaining the prediction accuracy around 50\%. 
To determine the optimal hint length, it is necessary to estimate the difficulty of the problem relative to the model capability at the current timestep. 
For this purpose, we introduce an accuracy estimator $f_\phi$ that captures the relationship between prediction accuracy and hint length 
(the specific form of $f_\phi$ and its optimization are introduced in Section~\ref{sec:irt}).

\begin{figure*}[t]
    \centering
    \includegraphics[width=0.98\linewidth]{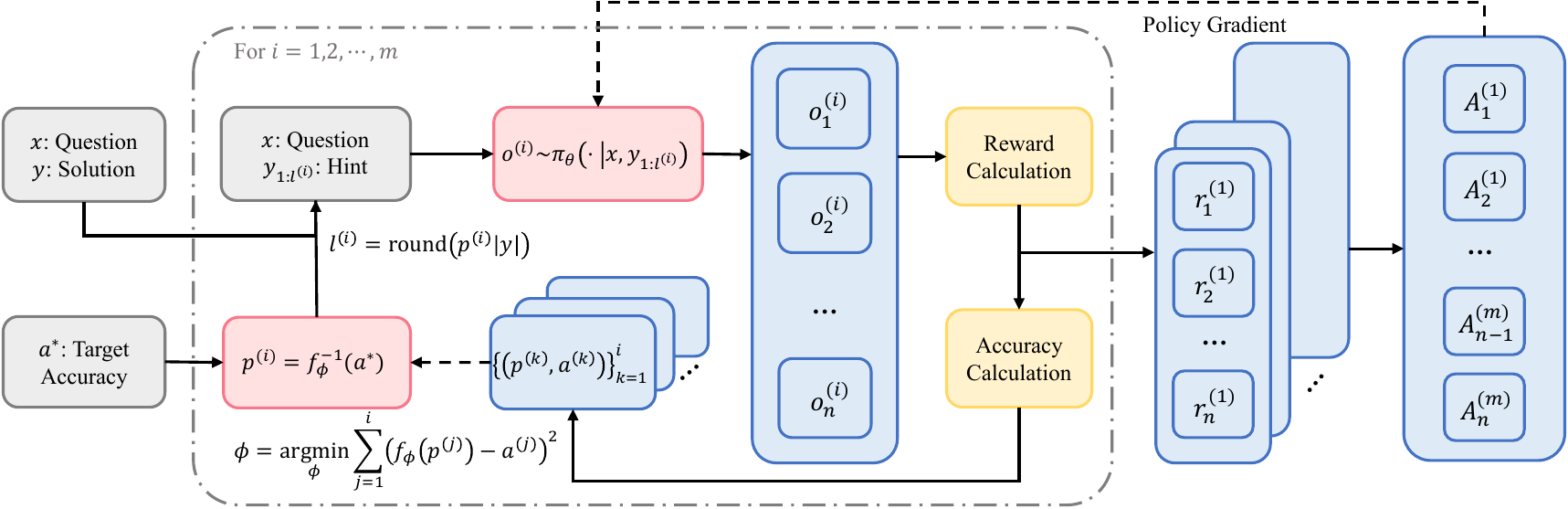}
    \caption{Overview of SEELE. In each step, SEELE conducts $m$ rollout rounds. 
    In round $i$, an adaptive hint $y_{1:l^{(i)}}$ is appended to the original problem, 
    where $y_{1:l^{(i)}}$ is a prefix of the full solution $y$ with length $l^{(i)}$, 
    determined by the accuracy-hint model $f_\phi$ to achieve the target rollout accuracy $a^*$. 
    The output accuracies within each round are then used to update $f_\phi$, enabling more accurate predictions in subsequent rounds. 
    Finally, SEELE computes the advantage function over the outputs of all $m$ rounds, which is used to update the policy.}
    \label{fig:overview}
\end{figure*}

As shown in Figure~\ref{fig:overview}, we design a multi-round adaptive sampling framework.
Different from standard GRPO rollout generation, SEELE distributes the rollouts into $m$ rounds to gradually fit the parameter $\phi$.
In round $i$, SEELE first predicts a hinting rate $p^{(i)}$ for reaching target accuracy $a^*$ by the inverse function of $f_\phi$.
Then, the corresponding part of the solution will be concatenated after $x$ as the input of the policy model for generating $n$ outputs ${o_1^{(i)}, o_2^{(i)},\cdots,o_n^{(i)}}$.
We calculate the accuracy within this round as $a^{(i)}$ and add the current hint-accuracy pair $(p^{(i)},a^{(i)})$ to the memory and update the parameter of $f_\phi$.
As the number of rounds increases, $f_\phi$ will model the accuracy more and more accurately and finally make the rollout accuracy stabilize at $a^*$.
Specifically, the first round needs a cold start where we use a default hinting rate $\frac{|y|-1}{|y|}$ in preparation for the worst model capability. The final hinting rate after $m$ rounds will be stored in the dataset so that in the next epoch SEELE can begin exploration from the last predicted rate.
Appendix~\ref{apsec:overhead} shows that the computational overhead of the multi-turn rollout is negligible.

After completing all rollout rounds, the $mn$ outputs will be used to calculate the advantages\footnote{Although the hints vary the prompts, the average baseline remains valid, as it is independent of the actions.}. In GRPO implementation, the advantage is computed at the response level and then distributed to all tokens, which will undermine the model's output probability on input hints if the model fails to explore a correct completion. Hence, we only compute RLVR loss on the generated tokens and impose a supervised loss on the hint tokens to encourage imitation. The final loss is
\begin{equation}\label{eq:seele_loss}
\Ls(\theta)
=-\E_{x\sim \train,\,o\sim\pi_\theta(\cdot|\hat x)}
    \big[
        A_{\theta_{\rm old}}(\hat x,o)
        +\gamma\log\pi_\theta(y_{1:l}|x)
    \big]
+ \beta\E_{x\sim \train}
    \big[
        \sD_{\rm KL}(\pi_{\theta_{\rm old}}(\cdot|\hat x)\,\|\,\pi_\theta(\cdot|\hat x))
    \big],
\end{equation}
where $\hat x$ denotes the concatenated input $x\oplus y_{1:l}$ and $y_{1:l}$ is the hint decided by multi-round sampling.

\subsection{Rollout Accuracy Prediction}\label{sec:irt}
The effectiveness of SEELE critically depends on the accuracy of the prediction model $f_\phi$, 
making its design particularly important. 
$f_\phi$ should be sufficiently expressive to fit the accuracy-hint relationship while also capable of generalizing from only a few data points. 
To construct such an accurate yet tractable model, we adopt an established framework used in humans.
In psychometrics, Item Response Theory (IRT)~\citep{chenItemResponseTheory2021} studies the relationship between an individual's performance on a test item and the test takers' levels of performance on an overall measure of the ability.
The three-parameter logistic (3PL) model is a widely used IRT model that gives the probability that a person with a given ability level will answer correctly:
\begin{equation}
    a(\Theta)=b+\frac{1-b}{1+e^{-k_0(\Theta-\nu)}},
\end{equation}
where $a$ and $\Theta$ indicate the test taker's prediction accuracy and capability, and three model parameters $\nu,k_0,b$ represent difficulty, discrimination, and guessing chance, respectively.

In our approach, the difficulty of the problem is tunable depending on the hinting rate. We assume the measure of difficulty $\nu$ is linearly related to the hinting rate $p$ as $\nu=-\frac{k}{k_0}p+\nu_0$. Through some symbol substitution and algebraic transformations, we derive the relationship between the accuracy and hinting rate at a certain level of model capability:
\begin{equation}\label{eq:irt}
    f_\phi(p)=b+\frac{1-b}{1+e^{-k(p+\mu)}},
\end{equation}
where $\phi$ represents the parameters $\{k,\mu,b\}$ and $\mu=\frac{k_0}{k}(\Theta-\nu_0)$ is the shifted capability measure.

The relationship described in Eq.~\eqref{eq:irt} aligns with both our intuition and empirical observations.  
When $p$ is small, the problem is too difficult for the model to explore a correct reasoning path, resulting in near-zero accuracy.  
As $p$ increases and critical steps are gradually revealed, the LLM becomes capable of completing the solution, leading to a rapid increase in accuracy.  
Once all critical steps are revealed, the model's accuracy approaches 1.0, and providing a longer hint yields no further gain.

Across different problems and training stages, the accuracy-hint curve varies in terms of its starting point, slope, and upper bound.  
We analyze these curves for 100 randomly sampled problems, with the model generating 100 outputs per problem at each hint level, and find that the 3PL model is sufficiently expressive to capture all observed patterns.  
Full illustrations are provided in Appendix~\ref{sec:acc_hint_full}, while Figure~\ref{fig:acc-hint} presents representative examples.  
The observed trends are consistent with IRT predictions, and the 3PL model accurately captures the relationship.  
Moreover, as the 3PL model involves only three parameters, it requires only a few rounds to obtain an accurate fit.
For the model fitting, SEELE employs non-linear least squares:
\begin{equation}\label{eq:phi}
    \phi=\argmin_{\phi}\sum_{j=1}^i\left(f_\phi(p^{(j)})-\hat a^{(j)}\right)^2,
\end{equation}
where $\hat a^{(j)}=\text{mean}(\{a^{(w)}|p^{(w)}\!=\!p^{(j)},w\!=\!1,\cdots,i\})$ is the averaged accuracy across the rounds with the same hinting rate, which helps reduce variance in the accuracy estimation.

\begin{figure}[t]
    \centering
    \includegraphics[width=0.99\linewidth]{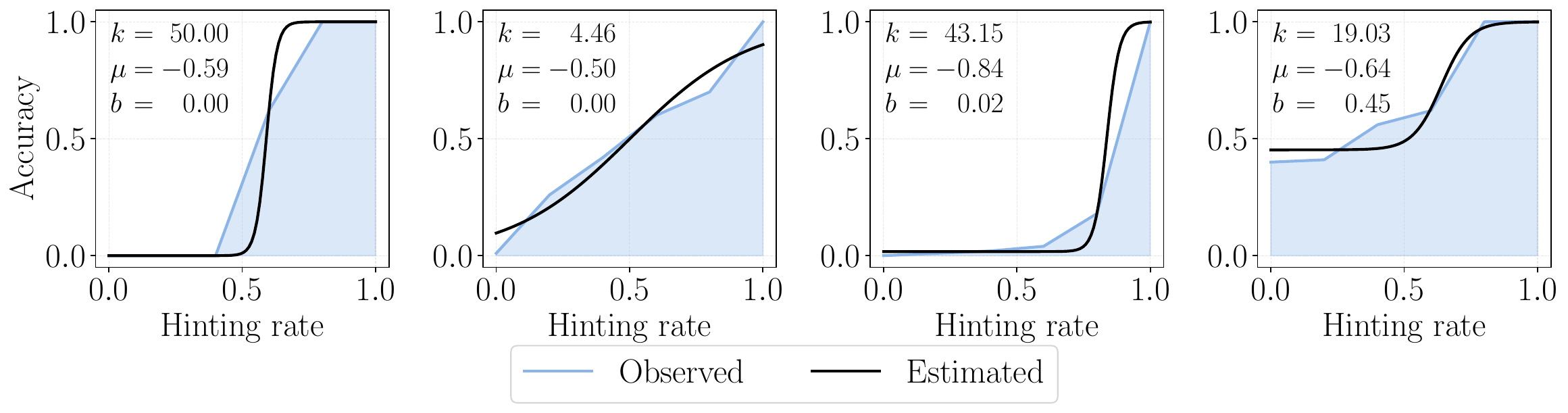}
    \caption{Cases of the accuracy-hint curves and the 3PL fitted curve and parameters. We select 4 typical curves to demonstrate the expressive power of the 3PL model.}
    \label{fig:acc-hint}
\end{figure}

\section{Experiments}

\subsection{Setup}\label{sec:setup}
\textbf{Training. }
We select DeepMath-103K as our training dataset. DeepMath-103K~\citep{heDeepMath103KLargeScaleChallenging2025} is a large-scale, decontaminated SFT dataset featuring challenging and verifiable mathematical problems, with a strong focus on higher-difficulty problems.
To construct a challenging training subset, we retain 22k problems that are particularly difficult, i.e., those on which Qwen2.5-7B~\citep{qwenQwen25TechnicalReport2025} fails to produce a correct answer.
For these problems, we annotate step-by-step reasoning traces using DeepSeek-V3~\citep{deepseek-aiDeepSeekV3TechnicalReport2025}.
Detailed data synthesis procedure and the annotation prompt are included in Appendix~\ref{apsec:data-synth}.
We adopt GRPO as the RL algorithm, setting the KL coefficient $\beta=0.001$ and the imitation coefficient $\gamma=0.001$.
We generate 32 rollouts in total with a maximum length of 2,048 tokens for each sample and set $m=4$ in our multi-round sampling.
The tested models include Qwen2.5-Math-7B, Qwen2.5-1.5B/3B, LLaMA 3.2-3B, and Mathstral-7B-v0.1.
More hyperparameter, algorithm and prompting details are included in Appendix~\ref{apsec:hyperparameter}, \ref{apsec:impl} and~\ref{apsec:cot-prompt}.

\textbf{Evaluation. }
Following prior works~\citep{zengSimpleRLZooInvestigatingTaming2025,huangBlendingSupervisedReinforcement2025}, we evaluate SEELE on 6 math and 3 general reasoning benchmarks.
The math benchmarks are GSM8K~\citep{cobbeTrainingVerifiersSolve2021}, MATH500~\citep{hendrycksMeasuringMathematicalProblem2021a}, Minerva~\citep{lewkowyczSolvingQuantitativeReasoning2022}, OlympiadBench~\citep{heOlympiadBenchChallengingBenchmark2024}, AIME24~\citep{numina_math_datasets}, and AMC23~\citep{aops_amc}.
The general benchmarks are ARC-Challenge~\citep{clarkThinkYouHave2018}, GPQA-Diamond~\citep{reinGPQAGraduateLevelGoogleProof2023}, and MMLU-Pro~\citep{wangMMLUProMoreRobust2024}.
For AIME24 and AMC23, we report avg@32 due to their small test sets and use a generation temperature of $0.6$; for others, we use greedy decoding and report pass@1 accuracy.

\textbf{Baselines. }
We compare SEELE with pure GRPO and SFT baselines and recent supervision-aided RLVR methods: LUFFY~\citep{yanLearningReasonOffPolicy2025}, UFT~\citep{liuUFTUnifyingSupervised2025}, and Prefix-RFT~\citep{huangBlendingSupervisedReinforcement2025}.
Among them, UFT and Prefix-RFT are concurrent hint-based approaches.
For a fair comparison, we train all the methods on the same dataset, prompt template, and models and set the hyperparameters following their papers.

\setlength\tabcolsep{2pt}
\begin{table*}[t]
    \centering
    \caption{Accuracy on math and general domain reasoning benchmarks.}\label{tab:main}
    \scriptsize
    \begin{tabular}{lccccccc|cccc}
        \toprule
        \multirow{2}{*}{\textbf{Model}} & \multicolumn{7}{c}{\textbf{Math Reasoning}} & \multicolumn{4}{c}{\textbf{General Domain Reasoning}}\\
        \cmidrule{2-8}\cmidrule{9-12}
        & \textbf{GSM8K} & \textbf{MATH500} & \textbf{Minerva} & \textbf{Olympiad} & \textbf{AIME24} & \textbf{AMC23} & \textbf{Avg.} & \textbf{ARC-C} & \textbf{GPQA-D} & \textbf{MMLU-Pro} & \textbf{Avg.}\\
        \midrule
        \textbf{Qwen2.5-Math-7B} & 71.6 & 63.2 & 25.7 & 32.0 & 14.1 & 45.2 & 42.0 & 69.5 & 24.7 & 17.7 & 37.3\\
        + SFT & 89.5 & 74.6 & 35.7 & 37.9 & 12.4 & 52.2 & 50.4 & 77.6 & 37.4 & 47.9 & 54.3\\
        + GRPO & 92.0 & 80.6 & 36.0 & 41.2 & 23.9 & 60.2 & 55.7 & 77.1 & 37.4 & 45.2 & 53.2\\
        + LUFFY & 91.7 & 80.0 & 35.3 & 42.4 & 18.5 & 66.2 & 55.7 & 80.5 & 39.9 & 49.9 & 56.8\\
        + UFT & 92.1 & 82.4 & 34.6 & 40.3 & 17.6 & 66.6 & 55.6 & \textbf{81.0} & 40.9 & 49.9 & 57.3\\
        + Prefix-RFT& 92.1 & 81.6 & 36.8 & 43.0 & 20.9 & 63.5 & 56.3 & 80.3 & 37.4 & 49.8 & 55.8\\
        \rowcolor{gray!10}+ SEELE (Ours)& \textbf{92.4} & \textbf{82.6} & \textbf{37.1} & \textbf{46.5} & \textbf{25.8} & \textbf{69.7} & \textbf{59.0} & 80.7 & \textbf{42.9} & \textbf{52.0} & \textbf{58.5}\\
        \midrule
        \textbf{Qwen2.5-1.5B} & 61.9 & 22.8 & 9.6 & 6.7 & 0.7 & 9.1 & 18.5 & 45.1 & 15.7 & 12.2 & 24.3\\
        + SFT & 67.4 & 43.6 & 13.6 & 12.6 & 1.4 & 16.4 & 25.8 & 63.7 & 25.8 & 30.2 & 39.9\\
        + GRPO & 70.1 & 36.4 & 10.7 & 11.1 & 1.8 & 15.2 & 24.2 & 64.8 & 25.3 & 23.1 & 37.7\\
        + LUFFY & 67.2 & 45.4 & 11.0 & 12.9 & 1.6 & 16.5 & 25.8 & 64.3 & 26.8 & 24.7 & 38.6\\
        + UFT & 72.6 & 50.4 & 12.9 & 15.9 & 3.9 & 26.6 & 30.4 & 66.1 & 23.7 & 28.5 & 39.4\\
        + Prefix-RFT& 71.5 & 48.0 & 13.6 & 14.5 & 2.1 & 22.7 & 28.7 & 65.3 & 23.7 & 25.0 & 38.0\\
        \rowcolor{gray!10}+ SEELE (Ours)& \textbf{76.5} & \textbf{58.0} & \textbf{16.2} & \textbf{19.9} & \textbf{4.1} & \textbf{30.4} & \textbf{34.2} & \textbf{68.3} & \textbf{27.8} & \textbf{31.7} & \textbf{42.6}\\
        \bottomrule
    \end{tabular}
\end{table*}

\subsection{Main results}\label{sec:main_results}
Table~\ref{tab:main} summarizes the performance on Qwen2.5-Math-7B and Qwen2.5-1.5B, while results on other models are included in Appendix~\ref{sec:more-model}.
SEELE consistently outperforms all baselines, demonstrating clear advantages across tasks and model scales.
For the larger Qwen2.5-Math-7B model, SEELE achieves the best math reasoning average of \textbf{59.0\%}, surpassing the strongest baseline (Prefix-RFT) by nearly 3 points, and also attains the highest general domain average of \textbf{58.5\%}.
As Qwen2.5-Math-7B is a well-trained math model, it benefits a lot from self-exploration and relies relatively less on the off-policy supervision, which accounts for GRPO's large advantage over SFT.
While LUFFY, UFT, and Prefix-RFT incorporate the supervision in the on-policy RL, their strategies are too rigid and fail to boost the training, only achieving marginal improvements over GRPO.

Compared with the 7B setting, SEELE shows even more pronounced advantages on the smaller Qwen2.5-1.5B model, where SEELE improves the math average to \textbf{34.2\%}, achieving an average improvement of \textbf{+10.0} points over GRPO, \textbf{+8.4} points over SFT, and \textbf{+3.8} points over the second best UFT.
On the general domain reasoning tasks, SEELE also outperforms the strongest baseline (SFT) by \textbf{+2.7} points.
The lower initial model capability of Qwen2.5-1.5B underscores the benefits of introducing off-policy demonstrations and leveraging dynamic data difficulty adaptation.
We further observe that GRPO generally underperforms SFT on complex reasoning tasks (e.g., MATH500, Minerva, AMC23), while showing comparable or slightly better performance on relatively easy tasks (e.g., GSM8K, ARC-C), demonstrating the limitations of purely on-policy exploration.
Moreover, the relatively low performance of GRPO and SFT indicates that exclusive self-exploration or pure imitation alone is insufficient to cultivate strong complex reasoning capabilities.

We provide additional pass@k metrics to assess generation diversity in Appendix~\ref{sec:passk}, as well as results demonstrating improvements on already strong reasoning models in Appendix~\ref{sec:strong}.

\begin{figure*}[t]
    \centering
    \includegraphics[width=0.98\linewidth]{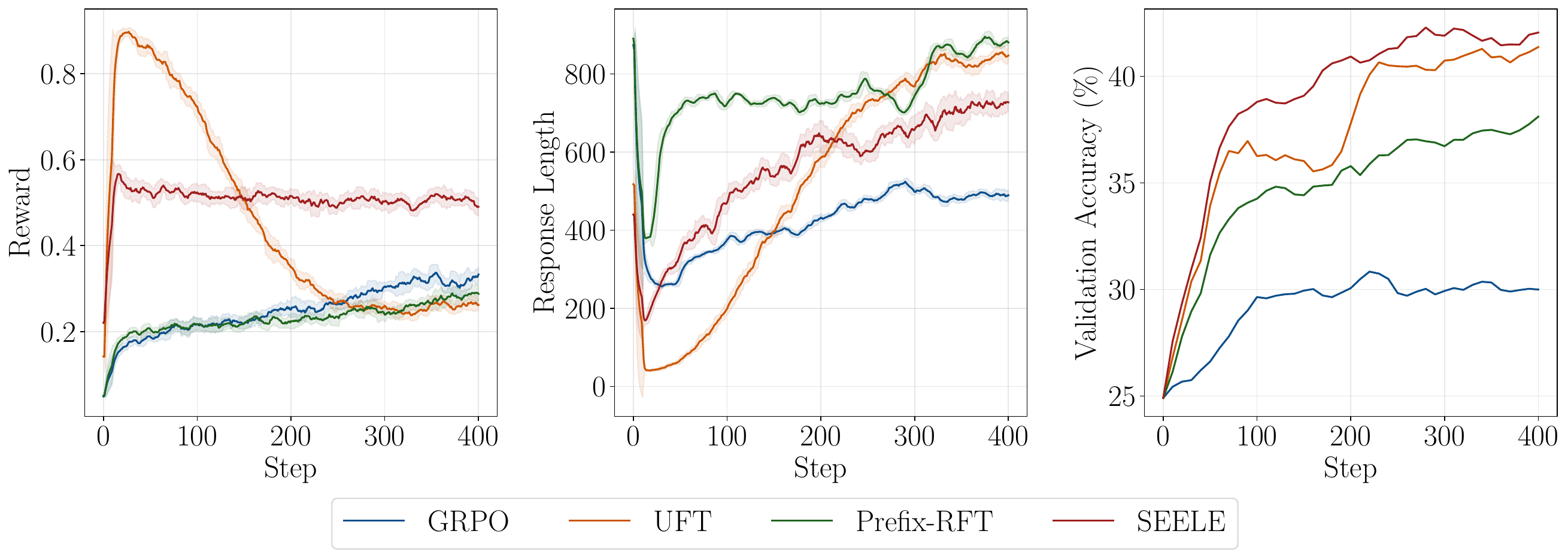}
    \caption{Training dynamics of RL compared with baselines on Qwen2.5-3B. \textbf{Left}: training rewards; \textbf{Middle}: Response length; \textbf{Right}: averaged accuracy on math reasoning validation sets.}
    \label{fig:dynamics}
\end{figure*}
\begin{figure}[t]
    \centering
    \includegraphics[width=0.95\linewidth]{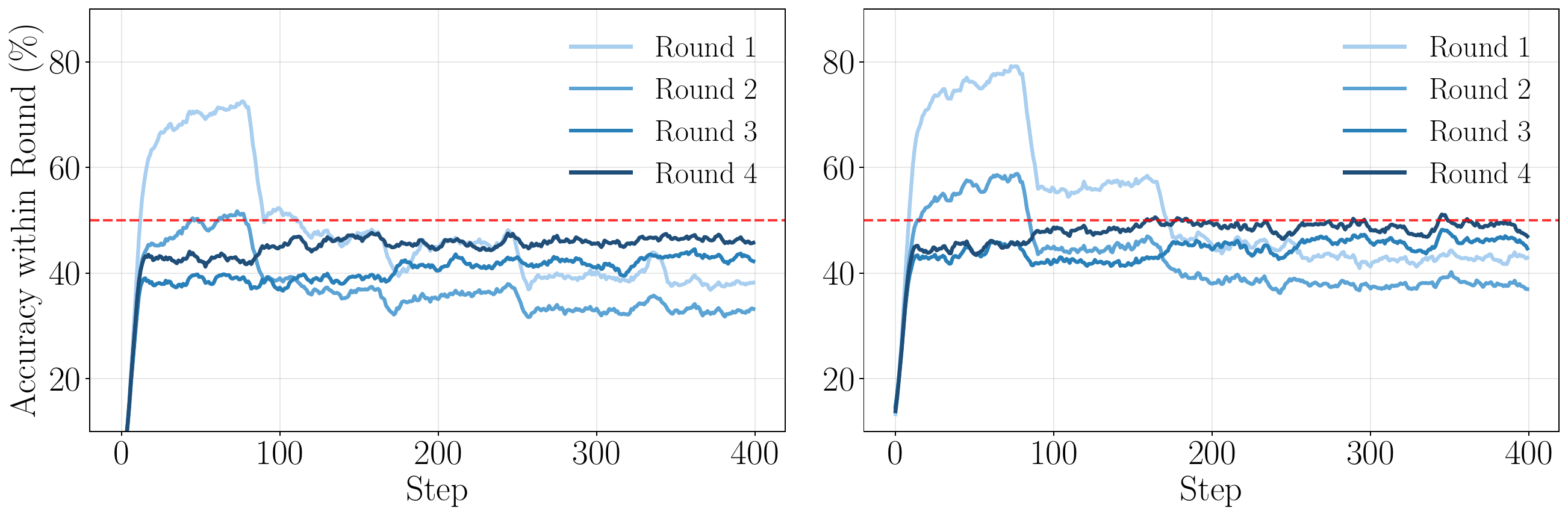}
    \caption{Accuracy within each rollout round during training. The red dotted line denotes the targeted accuracy. \textbf{Left}: Qwen2.5-1.5B; \textbf{Right}: Qwen2.5-3B.}
    \label{fig:round_acc}
\end{figure}

\subsection{Training Dynamics}

Figure~\ref{fig:dynamics} shows the training dynamics (reward, response length, and validation accuracy) of the SEELE, GRPO, and the two dynamic hinting baselines using Qwen2.5-3B.

\textbf{Precise Difficulty Control }
After the initial warm-up, the reward of SEELE rapidly rises to around 0.5 and maintains throughout the remaining training process, indicating our multi-round sampling and regression framework is able to precisely control the rollout accuracy.
Due to the adoption of a pessimistic cold-start strategy, the initial reward exceeds the target 0.5 and gradually decreases as training progresses.
To further understand SEELE's difficulty control mechanism, we visualize the intermediate results from the multi-round rollout sampling process in Figure~\ref{fig:round_acc}.
Relatively larger fluctuations and deviations are observed in Rounds~1 and~2, primarily because the 3PL regression model receives too few samples to produce accurate fits. 
Particularly, there is a peak in the first 80 steps (corresponding to the first epoch), which is caused by the high cold-start hinting rate.
From the second epoch onward, SEELE uses the hinting rate rectified in the preceding epoch for the initial round, resulting in more accurate predictions.
By Round~3 and Round~4, the accuracy is very close to the target 50\%, suggesting that the minimal data requirement for the 3PL model is approximately three samples, while four samples are sufficient to fit a model with adequate precision in practice.
We conduct further target control analysis in Appendix~\ref{sec:target_control} and demonstrate that controlling the accuracy to 50\% is beneficial.

\textbf{Accelerated Learning }
Compared with GRPO, UFT, and Prefix-RFT, SEELE consistently achieves higher accuracy throughout the entire training process and converges more rapidly. 
Within the first 100 steps, both SEELE and UFT quickly widen the accuracy gap over GRPO and Prefix-RFT, underscoring the importance of external guidance. 
GRPO reaches its performance ceiling around step~100, showing minimal improvement thereafter.
Notably, its reward continues to increase throughout the remainder of training and ultimately surpasses that of UFT and Prefix-RFT.
This behavior suggests that GRPO primarily reinforces previously acquired skills rather than facilitating the learning of new capabilities.
From step 100 to step 200, the performance of UFT stagnates because its problem difficulty fails to adapt to the model’s evolving capabilities.
UFT's holistic pre-designed decaying schedule cannot adequately meet the changing learning requirements.
In contrast, SEELE sustains a high growth rate until approximately step~300. 
From the perspective of response length, SEELE exhibits a more stable growth trend than UFT and Prefix-RFT, which aligns with the performance growth, highlighting the effectiveness of our hint adjustment schedule.

\subsection{Ablation Study on Hint-Length Modeling}

To verify the effectiveness of the proposed 3PL-based dynamic hinting strategy, we compare SEELE against fixed or random hint lengths $h \in \{1,2,3,4\}$ and a linear regression baseline ($a = kh + b$).

\begin{wraptable}{r}{0.38\columnwidth}
\centering
\caption{Ablation on hint-length modeling using Qwen2.5-1.5B.}
\label{tab:ablation_3pl}
\setlength{\tabcolsep}{5pt}
\begin{tabular}{lcc}
\toprule
Method & Math Avg & Gen Avg \\
\midrule
$h=1$   & 24.0 & 38.7 \\
$h=2$   & 31.5 & 40.7 \\
$h=3$   & 31.7 & 40.5 \\
$h=4$   & 31.5 & 38.6 \\
Random  & 32.1 & 38.4 \\
Linear  & 33.1 & 41.5 \\
3PL   & \textbf{34.2} & \textbf{42.6} \\
\bottomrule
\end{tabular}
\end{wraptable}

As shown in Table~\ref{tab:ablation_3pl}, selecting a fixed hint length yields poor performance across tasks. While moderate hint lengths ($h=2,3,4$) substantially outperform the minimal setting ($h=1$), no single static configuration consistently achieves the best results, indicating that the optimal hint ratio varies across problems.
Random hint lengths improve Math Avg over all fixed-length settings but underperform them on Gen Avg, suggesting that varying hint length alone is insufficient.
The linear regression baseline further improves over all fixed-length settings, demonstrating the benefit of dynamically adjusting hint length according to problem characteristics. This confirms that adaptive hinting is more effective than manually selecting a constant hint budget.

SEELE achieves the best performance on both math and general benchmarks, surpassing Linear by $+1.1$ points. These gains suggest that the relationship between hint ratio and learning effectiveness is inherently nonlinear, and therefore better captured by the proposed 3PL model than by a simple linear approximation. We provide further fitness analysis in Appendix~\ref{sec:acc_hint_full}.

\section{Conclusion}
In this paper, we present SEELE, a novel reinforcement learning with verifiable rewards (RLVR) framework that leverages off-policy demonstrations to dynamically align problem difficulty with the evolving capability of the model, thereby optimizing training efficiency.
This framework is motivated by our quantitative analysis, which shows that the upper bound on the learning efficiency of RL algorithms is maximized when the policy model's rollout accuracy is approximately 50\%.
To maintain rollout accuracy within this ``sweet spot'', we employ Item Response Theory and construct a multi-round regression model to predict the optimal hint length for fine-grained difficulty adjustment.
A key distinguishing feature of our approach, compared with previous supervision-aided RL methods, is that the difficulty manipulation operates at the instance level and incorporates real-time feedback, rendering the training process more responsive and enhancing the utility of each individual training sample.
Extensive experiments demonstrate that SEELE significantly outperforms GRPO, SFT, and other RLVR baselines.
Our study provides preliminary insights into the types of data favored by RL algorithms and offers a novel direction for improving data efficiency in reinforcement learning.

\section*{Limitations}
SEELE is primarily designed for scenarios where the difficulty of training problems exceeds the current capability of the model, such that adaptive hinting can effectively adjust the task difficulty into a more learnable regime. When the model is already sufficiently strong on the target tasks, the hinting mechanism may be activated less frequently or become unnecessary, in which case SEELE naturally degenerates to standard GRPO with limited additional benefit. In addition, SEELE relies on decomposing solutions into intermediate hints. For problems with very short reasoning chains or highly compressed solution processes, the granularity of hint segmentation becomes coarser, which may reduce the precision of difficulty control and weaken the advantage of our method. Exploring finer-grained hint construction and extending the framework to broader task settings remain important directions for future work.

\section*{Broader Impacts}
This work focuses on reinforcement learning with verifiable rewards for improving the reasoning performance of large language models, with experiments mainly conducted on mathematical and other structured tasks with objectively checkable answers. As such, we do not foresee significant new societal risks introduced specifically by our method beyond those already inherent to existing language models. Potential benefits include more accurate and reliable assistance in education, scientific research, and other applications that require verifiable reasoning. Our approach is algorithmic in nature and does not rely on sensitive personal data or deploy domain-specific decision systems.

\bibliography{reference,manual}
\bibliographystyle{plainnat}


\appendix

\section{Derivation of the RL Optimization Efficiency and Rollout Accuracy}\label{sec:derivation}
\subsection{One-step Unclipped PPO Objective}\label{sec:one-step-der}
We start from Eq.~\eqref{eq:obj} and assume the reference distribution $\pi_{\rm ref}$ equals $\pi_{\rm old}$.
Considering a one-step gradient update with the update vector $d$, we substitute $\theta=\theta_{\rm old}+d$ into the loss function and obtain the new loss value as
\begin{equation}
    \Ls(\theta_{\rm old}+d)=\Lpolicy(\theta_{\rm old}+d)+\beta\E_{x\sim \train}\left[\sD_{\rm KL}\left(\pi_{\theta_{\rm old}}(\cdot|x)||\pi_{\theta_{\rm old}+d}(\cdot|x)\right)\right].
\end{equation}
Note that $\Ls$ is defined as the independent sum of loss on each prompt $x$.
So, we can compute the individual loss for each $x$ and then aggregate the results.
For a specific $x$, we approximate $\Lpolicy(x;\theta_{\rm old}+d)$ and $\sD_{\rm KL}\left(\pi_{\theta_{\rm old}}(\cdot|x)||\pi_{\theta_{\rm old}+d}(\cdot|x)\right)$ using first-order and second-order Taylor expansion at $\theta_{\rm old}$, respectively.
First, we expand the policy loss:
\begin{equation}
    \Lpolicy(x;\theta_{\rm old}+d)\approx\Lpolicy(x;\theta_{\rm old})\\
    +\nabla_\theta\Lpolicy(x;\theta)\big|_{\theta=\theta_{\rm old}}^Td.
\end{equation}
\begin{equation}
    \begin{split}
    \sD_{\rm KL}\left(\pi_{\theta_{\rm old}}(\cdot|x)||\pi_{\theta_{\rm old}+d}(\cdot|x)\right)
    \approx&\sD_{\rm KL}\left(\pi_{\theta_{\rm old}}(\cdot|x)||\pi_{\theta_{\rm old}}(\cdot|x)\right)\\
    &\quad+\nabla_\theta\sD_{\rm KL}\left(\pi_{\theta_{\rm old}}(\cdot|x)||\pi_{\theta}(\cdot|x)\right)\big|_{\theta=\theta_{\rm old}}^Td\\
    &\quad+\frac{1}{2}d^T\nabla^2_\theta\sD_{\rm KL}\left(\pi_{\theta_{\rm old}}(\cdot|x)||\pi_{\theta}(\cdot|x)\right)\big|_{\theta=\theta_{\rm old}}d.
    \end{split}   
\end{equation}
Next, we compute the first- and second-order derivatives of KL-divergence.
\begin{align}
    \nabla_\theta\sD_{\rm KL}\left(\pi_{\theta_{\rm old}}(\cdot|x)||\pi_{\theta}(\cdot|x)\right)
    =&\nabla_\theta\E_{y\sim\pi_{\theta_{\rm old}}(\cdot|x)}\log\frac{\pi_{\theta_{\rm old}}(y|x)}{\pi_\theta(y|x)}\\
    =&-\nabla_\theta\E_{y\sim\pi_{\theta_{\rm old}}(\cdot|x)}\log{\pi_\theta(y|x)}\\
    =&-\E_{y\sim\pi_{\theta_{\rm old}}(\cdot|x)}\frac{\nabla_\theta\pi_\theta(y|x)}{\pi_\theta(y|x)}\\
    =&-\sum_y\frac{\pi_{\theta_{\rm old}}(y|x)}{\pi_\theta(y|x)}\nabla_\theta\pi_\theta(y|x).
\end{align}
\begin{align}
   \nabla_\theta\sD_{\rm KL}\left(\pi_{\theta_{\rm old}}(\cdot|x)||\pi_{\theta}(\cdot|x)\right)\big|_{\theta=\theta_{\rm old}}
   =&-\sum_y\frac{\pi_{\theta_{\rm old}}(y|x)}{\pi_{\theta_{\rm old}}(y|x)}\nabla_\theta\pi_\theta(y|x)\big|_{\theta=\theta_{\rm old}}\\
   =&-\nabla_\theta\left.\left[\sum_y\pi_\theta(y|x)\right]\right|_{\theta=\theta_{\rm old}}\\
   =&0.
\end{align}

\begin{align}
    \nabla^2_\theta\sD_{\rm KL}\left(\pi_{\theta_{\rm old}}(\cdot|x)||\pi_{\theta}(\cdot|x)\right)
    =&\nabla^2_\theta\E_{y\sim\pi_{\theta_{\rm old}}(\cdot|x)}\log\frac{\pi_{\theta_{\rm old}}(y|x)}{\pi_\theta(y|x)}\\
    =&-\nabla^2_\theta\E_{y\sim\pi_{\theta_{\rm old}}(\cdot|x)}\log{\pi_\theta(y|x)}\\
    =&-\E_{y\sim\pi_{\theta_{\rm old}}(\cdot|x)}\!\left[\frac{\nabla^2_\theta\pi_\theta(y|x)}{\pi_\theta(y|x)}-\frac{\nabla_\theta\pi_\theta(y|x)\nabla_\theta\pi_\theta(y|x)^T}{\pi^2_\theta(y|x)}\right]\\
    =&\E_{y\sim\pi_{\theta_{\rm old}}(\cdot|x)}\left[-\frac{\nabla^2_\theta\pi_\theta(y|x)}{\pi_\theta(y|x)}\!+\!\frac{\nabla_\theta\pi_\theta(y|x)}{\pi_\theta(y|x)}\frac{\nabla_\theta\pi_\theta(y|x)^T}{\pi_\theta(y|x)}\right]\\
    =&\E_{y\sim\pi_{\theta_{\rm old}}(\cdot|x)}\bigg[-\frac{\nabla^2_\theta\pi_\theta(y|x)}{\pi_\theta(y|x)}
    +\nabla_\theta\log\pi_\theta(y|x)\nabla_\theta\log\pi_\theta(y|x)^T\bigg]
\end{align}
We substitute $\theta=\theta_{\rm old}$ for the first term.
\begin{align}
    \E_{y\sim\pi_{\theta_{\rm old}}(\cdot|x)}\left.\left[\frac{\nabla^2_\theta\pi_\theta(y|x)}{\pi_\theta(y|x)}\right]\right|_{\theta=\theta_{\rm old}}
    &=\sum_y\frac{\pi_{\theta_{\rm old}}(y|x)}{\pi_{\theta_{\rm old}}(y|x)}\nabla^2_\theta\pi_\theta(y|x)\big|_{\theta=\theta_{\rm old}}\\
    &=\nabla^2_\theta\left.\left[\sum_y\pi_\theta(y|x)\right]\right|_{\theta=\theta_{\rm old}}\\
    &=0.
\end{align}
The second term is $F(\theta_{\rm old})$ when substituting $\theta=\theta_{\rm old}$.

So, we get
\begin{equation}\label{eq:ap-approx}
    \Ls(x;\theta_{\rm old}+d)\approx\Lpolicy(x;\theta_{\rm old})
    +\nabla_\theta\Lpolicy(x;\theta)\big|_{\theta=\theta_{\rm old}}^Td+\frac{\beta}{2}d^TF(\theta_{\rm old})d.
\end{equation}
Since the Fisher information matrix is positive semi-definite, the right-hand side of Eq.~\eqref{eq:ap-approx} is convex and has a global minimizer. To find the minimizer $d_x^*$, let the derivative be zero.
\begin{equation}
    \nabla_\theta\Lpolicy(x;\theta)\big|_{\theta=\theta_{\rm old}}+\beta F(\theta_{\rm old})d_x^*=0.
\end{equation}
\begin{equation}
    d_x^*=-\frac{1}{\beta}F(\theta_{\rm old})^{-1}\nabla_\theta\Lpolicy(x;\theta)\big|_{\theta=\theta_{\rm old}}.
\end{equation}
Substitute back to Eq.~\eqref{eq:ap-approx}
\begin{equation}
    \Ls(x;\theta_{\rm old})-\Ls(x;\theta_{\rm old}+d_x^*)
    \approx\frac{1}{2\beta}\nabla_\theta\Lpolicy(x;\theta)\big|_{\theta=\theta_{\rm old}}^TF^{-1}(\theta_{\rm old})\nabla_\theta\Lpolicy(x;\theta)\big|_{\theta=\theta_{\rm old}}.
\end{equation}
Now, we need to connect $\nabla_\theta\Lpolicy(x;\theta)\big|_{\theta=\theta_{\rm old}}$ with $a_{\theta_{\rm old}}(x)$. According to the definition
\begin{align}
    \nabla_\theta\Lpolicy(x;\theta)
    =&-\nabla_\theta\E_{y\sim\pi_{\theta}(\cdot|x)}A_{\theta_{\rm old}}(x,y)\\
    =&-\nabla_\theta\E_{y\sim\pi_{\theta}(\cdot|x)}\left[r(x,y)-\E_{y\sim\pi_{\theta_{\rm old}}(\cdot|x)}r(x,y)\right]\\
    =&-\nabla_\theta\E_{y\sim\pi_{\theta}(\cdot|x)}\left[r(x,y)\right]\\
    =&-\nabla_\theta a_{\theta}(x).
\end{align}
\begin{equation}\label{eq:ap-aFa}
    \begin{split}
    \Ls(x;\theta_{\rm old})-\Ls(x;\theta_{\rm old}+d_x^*)
    \approx&\frac{1}{2\beta}\nabla_\theta a_{\theta}(x)\big|_{\theta=\theta_{\rm old}}^T
    F^{-1}(\theta_{\rm old})\nabla_\theta a_{\theta}(x)\big|_{\theta=\theta_{\rm old}}.
    \end{split}
\end{equation}
Note that $r(x,y)$ is an unbiased estimator of $a_\theta(x)$.
By applying the vector parameter Cramér–Rao bound~\citep[Section~3.8]{kay1993fundamentals} to Eq.~\eqref{eq:ap-aFa}, we get
\begin{align}
    \Ls(x;\theta_{\rm old})-\Ls(x;\theta_{\rm old}+d_x^*)
    \leq&\frac{1}{2\beta}\Var_{y\sim\pi_{\theta_{\rm old}}(\cdot|x)}[r(x,y)]\\
    =&\frac{1}{2\beta}a_{\theta_{\rm old}}(x)(1-a_{\theta_{\rm old}}(x)).
\end{align}
So, for an arbitrary $d$
\begin{equation}
    \Ls(x;\theta_{\rm old})-\Ls(x;\theta_{\rm old}+d)\leq\frac{1}{2\beta}a_{\theta_{\rm old}}(x)(1-a_{\theta_{\rm old}}(x)).
\end{equation}
Now we have got the inequality for a single $x$, taking the expectation we get
\begin{equation}~\label{eq:one-step-final}
    \Ls(\theta_{\rm old})-\Ls(\theta_{\rm old}+d)\leq\frac{1}{2\beta}\E_{x\sim \train}[a_{\theta_{\rm old}}(x)(1-a_{\theta_{\rm old}}(x))].
\end{equation}

\subsection{Multi-Step Clipped PPO Objective}
\label{sec:multistep_extension}

Training performs $K$ updates using different mini-batches from a fixed rollout batch, while keeping the sampling policy $\pi_{\rm old}=\pi_{\theta_{\rm old}}$ and the advantages fixed. We take $\pi_{\rm ref}=\pi_{\rm old}$. To examine how the accuracy affects the decrease in the clipped loss, we analyze a local, per-prompt model of these updates.
The proof proceeds in three steps. First, we bound the parameter displacement after $K$ updates. Second, we combine this displacement bound with an upper bound on the clipped advantage and a lower bound on the KL cost, obtaining a scalar maximization problem that bounds the loss decrease. Third, we derive a closed-form upper bound on this maximum.

\paragraph{Preliminaries: Local loss and update rule.}
Fix a prompt $x$. Write $a=a_{\theta_{\rm old}}(x)$ and $v=a(1-a)$, and abbreviate the binary reward advantage $A_{\theta_{\rm old}}(x,y)$ by $A=r(x,y)-a$. Let $\E_{\rm old}$ denote expectation over $y\sim\pi_{\rm old}(\cdot|x)$ and $\rho=\pi_\theta(y|x)/\pi_{\rm old}(y|x)$. Omitting the fixed argument $x$ inside KL divergences, the loss is
\begin{equation}\label{eq:a2-objective}
\Ls^{\rm clip}(x;\theta)=-\E_{\rm old}\!\left[\min\{\rho A,\operatorname{clip}(\rho,1-\epsilon,1+\epsilon)A\}\right]+\beta\,\sD_{\rm KL}(\pi_{\rm old}\|\pi_\theta).
\end{equation}
Here $\beta>0$ and $0<\epsilon<1$. Define the score, Fisher matrix, and initial policy gradient by
\begin{equation}\label{eq:a2-score}
s(y)=\left.\nabla_\theta\log\pi_\theta(y|x)\right|_{\theta=\theta_{\rm old}},
\qquad F=\E_{\rm old}[s(y)s(y)^T],
\qquad g=\E_{\rm old}[A s(y)].
\end{equation}
We assume locally fixed policy support, twice differentiable probabilities, valid differentiation under the expectations, and finite second moments of $s$. Then $\E_{\rm old}[s]=0$, $g$ is the accuracy gradient at $\theta_{\rm old}$, and $F$ is the Hessian of the KL term there.

We first express the loss in terms of the displacement $d=\theta-\theta_{\rm old}$. This requires the following local approximation.

\noindent\textbf{Assumption 1 (Local approximation).}
All $K$ updates remain close enough to $\theta_{\rm old}$ for the first-order expansion of the policy ratio and the second-order expansion of the KL term to remain accurate:
\begin{equation}\label{eq:a2-local-approximations}
\rho\approx 1+s(y)^Td,\qquad
\sD_{\rm KL}(\pi_{\rm old}\|\pi_{\theta_{\rm old}+d})\approx \frac12 d^TFd.
\end{equation}
The identity $\min\{\rho A,\operatorname{clip}(\rho,1-\epsilon,1+\epsilon)A\}=A+\min\{A(\rho-1),\epsilon|A|\}$ and $\E_{\rm old}[A]=0$ give the local loss
\begin{equation}\label{eq:a2-local-loss}
\widetilde{\Ls}^{\rm clip}(x;d)=-J(d)+\frac{\beta}{2}d^TFd,
\qquad J(d)=\E_{\rm old}\!\left[\min\{A s(y)^Td,\epsilon|A|\}\right].
\end{equation}
Thus $\Ls^{\rm clip}(x;\theta_{\rm old}+d)\approx\widetilde{\Ls}^{\rm clip}(x;d)$, with the clipping operation retained. To obtain the update rule, we first differentiate this local loss. A response contributes $A s(y)$ to the gradient of $J$ while its contribution is on the linear branch, and zero when it is on the constant, clipped branch. Hence, at $d_k=\theta_k-\theta_{\rm old}$, define
\begin{equation}\label{eq:a2-policy-gradient}
g_k=\E_{\rm old}[\chi_k A s],\qquad
\nabla_d\widetilde{\Ls}^{\rm clip}(x;d_k)=-g_k+\beta Fd_k.
\end{equation}
Here $\chi_k(y)=1$ if $A s(y)^Td_k<\epsilon|A|$ and $\chi_k(y)=0$ if the inequality is reversed. At equality, any value in $[0,1]$ gives a valid subgradient in place of the gradient above.

Having obtained the loss gradient, we next specify how it changes the parameters.

\noindent\textbf{Assumption 2 (Gradient updates).}
We represent mini-batch gradients by their expectations under $\pi_{\rm old}$ and apply gradient descent with step size $\eta>0$, starting from $d_0=0$.

Since $\theta_{\rm old}$ is fixed, updating $\theta$ is equivalent to updating $d$. Substituting Eq.~\eqref{eq:a2-policy-gradient} into the gradient descent step gives
\begin{equation}\label{eq:a2-update}
d_{k+1}=d_k-\eta(-g_k+\beta Fd_k)=(I-\eta\beta F)d_k+\eta g_k.
\end{equation}
At each step, we recompute $g_k$ according to which responses are clipped under the current parameters.

Because the local KL cost is $\beta d^TFd/2$, we measure displacement by $\|d\|_F=(d^TFd)^{1/2}$. The local loss starts at zero, so its decrease after $K$ steps is
\begin{equation}\label{eq:a2-decrease}
\Delta_K(x):=\widetilde{\Ls}^{\rm clip}(x;0)-\widetilde{\Ls}^{\rm clip}(x;d_K)
=J(d_K)-\frac{\beta}{2}\|d_K\|_F^2.
\end{equation}
We first bound the displacement reached by Eq.~\eqref{eq:a2-update}, then combine an upper bound on the clipped gain with a lower bound on the KL cost. The following inequalities are exact for the local model under the stated conditions.

\paragraph{Step 1: Bounding the parameter displacement.}
Equation~\eqref{eq:a2-update} contains two terms to control: the new gradient contribution $\eta g_k$ and the previous displacement multiplied by $I-\eta\beta F$. We start with the gradient. Since $r$ is binary, $\E_{\rm old}[A^2]=a(1-a)=v$. For any direction $w$, Cauchy--Schwarz and $0\le\chi_k\le1$ give
\begin{equation}\label{eq:a2-directional-bound}
(w^Tg_k)^2\le\E_{\rm old}[\chi_k^2A^2]\,\E_{\rm old}[(w^Ts)^2]\le v\,w^TFw.
\end{equation}
Taking $w=F^{-1}g_k$\footnote{If $F$ has zero eigenvalues, its inverse is understood on the subspace spanned by eigenvectors with positive eigenvalues. Indeed, $Fw=0$ implies $\E_{\rm old}[(w^Ts)^2]=0$, hence $w^Tg_k=0$. Starting from $d_0=0$, the updates have no component in these zero-eigenvalue directions. If $F=0$, all gradients and the loss decrease are zero.} yields
\begin{equation}\label{eq:a2-gradient-scale}
\|g_k\|_{F^{-1}}^2=g_k^TF^{-1}g_k\le v.
\end{equation}
To use the gradient bound in the displacement recursion, we still need to convert it from the $F^{-1}$-norm to the $F$-norm. We also need to control how $I-\eta\beta F$ changes the previous displacement. Both depend on the eigenvalues of $F$. In particular, along an eigenvector with eigenvalue $\lambda$, this matrix multiplies that component by $1-\eta\beta\lambda$. We therefore introduce the following bounds.

\noindent\textbf{Assumption 3 (Curvature and step-size bounds).}
Let $\mu>0$ and $M\ge\mu$ be lower and upper bounds on the positive eigenvalues of $F$, respectively. We assume
\begin{equation}\label{eq:a2-curvature}
\mu\le\lambda\le M\quad\text{for every positive eigenvalue $\lambda$ of $F$},
\qquad 0<\eta\beta M\le1.
\end{equation}
The step-size condition keeps $1-\eta\beta\lambda$ nonnegative, while the lower curvature bound makes it at most $q:=1-\eta\beta\mu<1$. Consequently, $\|(I-\eta\beta F)d\|_F\le q\|d\|_F$. The upper curvature bound also gives $\|g_k\|_F\le M\|g_k\|_{F^{-1}}\le M\sqrt v$. Applying these two bounds to Eq.~\eqref{eq:a2-update} yields
\begin{equation}\label{eq:a2-radius-recursion}
\|d_{k+1}\|_F\le q\|d_k\|_F+\eta M\sqrt v.
\end{equation}
Starting from $d_0=0$ and summing the geometric series, we obtain
\begin{equation}\label{eq:a2-radius}
\|d_K\|_F\le\sqrt v\,h_K,\qquad
h_K=\eta M\sum_{j=0}^{K-1}q^j
=\frac{M}{\beta\mu}\left[1-(1-\eta\beta\mu)^K\right].
\end{equation}
This bound tracks the accumulation of successive clipped gradients together with the contraction caused by the KL term.

\paragraph{Step 2: Accounting for clipping and the KL cost.}
We now bound the clipped gain $J(d)$ within the displacement bound. Since binary rewards give two advantage values, we treat correct and incorrect responses separately. For $0<a<1$, set $m=\min(a,1-a)$ and $z=g^Td/v$. Here $vz=g^Td$ is the first-order change in the prompt's accuracy. Using $\E_{\rm old}[s]=0$, we have $g^Td=\E_{\rm old}[r\,s^Td]=a\,\E_{\rm old}[s^Td\mid r=1]$. Combining this with the zero mean of $s^Td$ gives
\begin{equation}\label{eq:a2-class-means}
\E_{\rm old}[s^Td\mid r=1]=(1-a)z,\qquad
\E_{\rm old}[-s^Td\mid r=0]=az.
\end{equation}
For correct and incorrect responses, the clipped contributions to $J(d)$ are $(1-a)\min\{s^Td,\epsilon\}$ and $a\min\{-s^Td,\epsilon\}$, respectively. Weighting the two classes by their probabilities $a$ and $1-a$ gives
\[
J(d)=v\!\left[\E_{\rm old}[\min\{s^Td,\epsilon\}\mid r=1]+\E_{\rm old}[\min\{-s^Td,\epsilon\}\mid r=0]\right].
\]
Since $u\mapsto\min(u,\epsilon)$ is concave, we apply Jensen's inequality to $s^Td$ conditional on $r=1$ and to $-s^Td$ conditional on $r=0$. Using the conditional means derived above yields
\begin{align}
\E_{\rm old}[\min\{s^Td,\epsilon\}\mid r=1]
&\le \min\{\E_{\rm old}[s^Td\mid r=1],\epsilon\}
=\min\{(1-a)z,\epsilon\},\\
\E_{\rm old}[\min\{-s^Td,\epsilon\}\mid r=0]
&\le \min\{\E_{\rm old}[-s^Td\mid r=0],\epsilon\}
=\min\{az,\epsilon\}.
\end{align}
Substituting these two bounds into the expression for $J(d)$ gives
\begin{equation}\label{eq:a2-clipping-envelope}
J(d)\le v\!\left[\min\{(1-a)z,\epsilon\}+\min\{az,\epsilon\}\right]
=vS_a(z),\qquad S_a(z)=\min\{z,\epsilon+mz,2\epsilon\}.
\end{equation}
For $z\ge0$, the slope of $S_a$ decreases from $1$ to $m$ and then to $0$, at $z=\epsilon/(1-m)$ and $z=\epsilon/m$. The first change occurs when one class reaches its clipping limit in this bound; the second occurs when both do. At $a=1/2$, the thresholds coincide. These are thresholds of the bound on the expected contribution; individual responses can become clipped at different steps.

By Eq.~\eqref{eq:a2-decrease}, the loss decrease is $\Delta_K(x)=J(d_K)-\frac{\beta}{2}d_K^TFd_K$. We have bounded the policy gain from above in terms of $z=g^Td/v$. We next express a lower bound on the KL cost in terms of the same variable, so that the two bounds can be combined.

At the initial point $d_0=0$, clipping is inactive and hence $g_0=g$. Applying Eq.~\eqref{eq:a2-directional-bound} with $k=0$ and $w=d$ gives $(g^Td)^2\le v\,d^TFd$. Since $g^Td=vz$, this implies $d^TFd\ge vz^2$. Combining this lower bound with Eq.~\eqref{eq:a2-clipping-envelope} at $d=d_K$, where $z_K=g^Td_K/v$, yields
\[
\Delta_K(x)=J(d_K)-\frac{\beta}{2}d_K^TFd_K
\le v\left[S_a(z_K)-\frac{\beta}{2}z_K^2\right].
\]

This bound still depends on the value of $z_K$ reached after $K$ updates. To bound its range, Eq.~\eqref{eq:a2-radius} gives $d_K^TFd_K\le vh_K^2$. Together with the inequality above, we obtain
\[
v^2z_K^2=(g^Td_K)^2\le v\,d_K^TFd_K\le v^2h_K^2,
\]
and therefore $|z_K|\le h_K$. We can thus bound the loss decrease by maximizing $v[S_a(z)-\beta z^2/2]$ over $[-h_K,h_K]$. For $z<0$, we have $S_a(z)=z$, so this expression is negative, whereas its value at $z=0$ is zero. The maximum can therefore be restricted to $[0,h_K]$, giving
\begin{equation}\label{eq:a2-scalar-bound}
\Delta_K(x)\le v\max_{0\le z\le h_K}\left[S_a(z)-\frac{\beta}{2}z^2\right].
\end{equation}

\paragraph{Step 3: Derivation of the bound.}
Write $f_a(z)=S_a(z)-\beta z^2/2$. Since all slopes of $S_a$ are at most $1$, for $z\ge1/\beta$,
\begin{equation}
f_a(z)-f_a(1/\beta)
\le z-\frac1\beta-\frac{\beta}{2}
\left(z^2-\frac1{\beta^2}\right)
=-\frac{\beta}{2}\left(z-\frac1\beta\right)^2
\le0.
\end{equation}
Thus, the maximum over $[0,h_K]$ equals that over
$[0,\min\{h_K,1/\beta\}]$. Define
\begin{equation}\label{eq:a2-tK}
t_K=\min\{1,\beta h_K\}
=\min\!\left\{1,\frac{M}{\mu}
\left[1-(1-\eta\beta\mu)^K\right]\right\}.
\end{equation}
The maximization interval is therefore $[0,t_K/\beta]$.

To evaluate an upper bound on this maximum, use
$S_a(z)\le\min\{z,2\epsilon\}$ and
$S_a(z)\le\epsilon+mz$. Define
\begin{align}
U_K&:=\max_{0\le z\le t_K/\beta}
\left[\min\{z,2\epsilon\}-\frac{\beta}{2}z^2\right],\\
V_K&:=\max_{0\le z\le t_K/\beta}
\left[\epsilon+mz-\frac{\beta}{2}z^2\right].
\end{align}
Both expressions bound the same maximum from above, so
\begin{equation}
\Delta_K(x)
\le v\max_{0\le z\le t_K/\beta}f_a(z)
\le v\min\{U_K,V_K\}.
\end{equation}
Since $t_K\le1$, the maximizers defining $U_K$ and $V_K$ are, respectively,
\begin{equation}
z_1^*=\min\{t_K/\beta,2\epsilon\},
\qquad
z_2^*=\min\{t_K,m\}/\beta.
\end{equation}
Substituting these maximizers gives the following conclusion.

\paragraph{Final bound.}
With $(u)_+=\max\{u,0\}$ and
\begin{equation}\label{eq:a2-UK}
U_K=
\frac{1-\max\{1-t_K,(1-2\beta\epsilon)_+\}^{2}}{2\beta},
\end{equation}
the loss decrease after $K$ updates satisfies
\begin{equation}\label{eq:a2-bound}
\Delta_K(x)\le B_K(a)
:=a(1-a)\min\!\left\{
U_K,\;
\epsilon+\frac{m^2-(m-t_K)_+^2}{2\beta}
\right\}.
\end{equation}
For batch updates, we assume that the common displacement $d_K$ satisfies the radius condition in Eq.~\eqref{eq:a2-radius} for every prompt. The preceding pointwise bound then applies along this shared trajectory, and averaging gives
\begin{equation}\label{eq:total}
\E_{x\sim\train}[\Delta_K(x)]
\le \E_{x\sim\train}\!\left[
B_K\!\left(a_{\theta_{\rm old}}(x)\right)
\right].
\end{equation}

\paragraph{Interpretation.}
The factor $t_K$ captures the cumulative effect of $K$ updates under the specified step size, KL strength, and curvature bounds. The term $U_K$ combines the clipping limit $2\epsilon$ with the quadratic KL cost.
We assume common curvature bounds $\mu,M$ and fixed hyperparameters $\eta,\beta,\epsilon$ across prompts, so that $t_K$ and $U_K$ are independent of $a$. The accuracy dependence of $B_K(a)$ is then entirely through $m=\min\{a,1-a\}$. As $m$ increases from $0$ to $1/2$, the factor $a(1-a)=m(1-m)$ strictly increases, while $m^2-(m-t_K)_+^2$ is nondecreasing. For $K\ge1$, the second factor in Eq.~\eqref{eq:a2-bound} is therefore positive and nondecreasing, making $B_K(a)$ strictly increasing in $m$. Thus, $B_K(a)$ is symmetric about $a=1/2$, vanishes at $a\in\{0,1\}$, and attains its unique maximum at $a=1/2$.
The bound retains the sweet spot at $a=1/2$ while accounting for repeated updates, clipping, and KL regularization.

\begin{algorithm}[t]
    \caption{SEELE Policy Optimization}\label{algo:seele}
    \begin{algorithmic}[1]
        \REQUIRE initial policy model $\pi_{\rm init}$; training set $\train$; initial hint rate $p_0$
        \STATE policy model $\pi_\theta\gets\pi_{\rm init}$
        \STATE reference model $\pi_{\rm ref}\gets\pi_{\rm init}$
        \FOR{step=$1,2,\cdots,T$}
            \STATE $\pi_{\theta_{\rm old}}\gets\pi_\theta$
            \STATE Sample a batch $B$ from $\train$
            \STATE $C\gets\{x:\emptyset\}_{x\in B}$\quad(Here $C_x$ refers to $\{(p_x^{(j)},a_x^{(j)})\}_{j=1}^{i-1}$)
            \FOR{$i=1,2,\cdots,m$}
                \STATE $B_i\gets\emptyset$
                \FOR{$x,y\in B$}
                    \IF{$i == 1$}
                        \STATE $p_x^{(i)}\gets p_0$ if epoch $==1$, else $p_x^{(m)}$ from the previous epoch
                    \ELSE
                    \STATE $\hat C_x\gets C_x\cup\{(w,w)|w\in\{0,1\}\text{\textbackslash}\{p_x^{(j)}\}_{j=1}^{i-1}\}$
                    \STATE Optimize $\phi$ on $\hat C_x$ (Follow Eq.~\eqref{eq:phi}.)
                    \STATE $p_x^{(i)}=f^{-1}_\phi(a^*)$
                    \ENDIF
                    \STATE $\hat x=x\oplus y_{1:l},\quad l={\rm round}(p_x^{(i)}|y|)$
                    \STATE $B_i\gets B_i\cup\{(\hat x,y)\}$
                \ENDFOR
                \STATE Sample $n$ outputs $\{o_j^{(i)}\}_{j=1}^n\sim\pi_{\theta_{\rm old}}(\cdot|\hat x)$ for each $\hat x\in B_i$
                \STATE Compute rewards $\{r_j^{(i)}\}_{j=1}^n$ for each sampled output
                \STATE Compute rollout accuracy $a_x^{(i)}$ over $\{r_j^{(i)}\}_{j=1}^n$ for each problem $x$
                \FOR{$x\in B$}
                    \STATE $C_x\gets C_x\cup\{(p_x^{(i)},a_x^{(i)})\}$
                \ENDFOR
            \ENDFOR
            \STATE Compute advantage $\{A_{j,t}^{(i)}\}_{i=1,j=1}^{m,n} $ for each sampled token
            \FOR{GRPO iteration=$1,2\cdots,u$}
                \STATE Update the policy model $\pi_\theta$ by minimizing the objective in Eq.~\eqref{eq:impl-loss}.
            \ENDFOR
        \ENDFOR
        \ENSURE $\pi_\theta$
    \end{algorithmic}
\end{algorithm}

\section{Training Details}\label{apsec:training_details}
\subsection{Hyperparameter Setting}\label{apsec:hyperparameter}
We adopt GRPO as the RL algorithm, setting the KL coefficient $\beta=0.001$ and the imitation coefficient $\gamma=0.001$.
Our rollout batch size is 256 and the update batch size is 64.
For our approach and all other RL-based baselines, we generate 32 rollouts in total with a maximum length of 2,048 tokens for each sample.
For our multi-round sampling, we set the number of rounds $m=4$ and each round will generate $n=8$ rollouts.
Temperature is set to 1.0 for the rollout generation.
Our training is based on veRL~\citep{shengHybridFlowFlexibleEfficient2025}.
We use MathRuler~\citep{mathruler} to verify the correctness of the model's outputs and use the TRF non-linear least squares algorithm provided by the LMFIT library~\citep{lmfit} to fit $f_\phi$.
Our experiments are conducted under the Zero-RL setting, where we use the DeepSeek-R1-Zero prompting template and train the base model for 400 steps. We train SEELE for 2 days on the 7B model using 8 A100 GPUs.

\subsection{Policy Optimization Implementation Details}\label{apsec:impl}
In the practical implementation, we apply the clipping techniques used in GRPO~\citep{shaoDeepSeekMathPushingLimits2024} and use $\pi_{ref}$ rather than $\pi_{\rm old}$ in KL regularization for simplicity, as shown in Eq.~\eqref{eq:impl-adv} and Eq.~\eqref{eq:impl-loss}, which slightly differs from the formulation in Eq.~\eqref{eq:seele_loss}.
\begin{equation}\label{eq:impl-loss}
    \Ls(\theta)=-\E_{x\sim \train,o\sim\pi_{\theta_{\rm old}}(\cdot|\hat x)}[\hat A_{\theta_{\rm old}}(\hat x,o)+\gamma\log\pi_\theta(y_{1:l}|x)]
    +\beta\E_{x\sim \train}[\sD_{\rm KL}(\pi_{\theta_{\rm ref}}(\cdot|\hat x)||\pi_\theta(\cdot|\hat x))],
\end{equation}
\begin{equation}\label{eq:impl-adv}
    \hat A_{\theta_{\rm old}}(\hat x,o)=\sum_{t=1}^{|o|}\bigg\{\min\bigg[\frac{\pi_\theta(o_t|\hat x,o_{<t})}{\pi_{\theta_{\rm old}}(o_t|\hat x,o_{<t})}A_t,
    {\rm clip}\left(\frac{\pi_\theta(o_t|\hat x,o_{<t})}{\pi_{\theta_{\rm old}}(o_t|\hat x,o_{<t})},1-\epsilon,1+\epsilon\right)A_t\bigg]\bigg\}
\end{equation}
In Algorithm~\ref{algo:seele}, we present the detailed workflow of SEELE policy optimization.  
At each training step, SEELE first samples a batch from the training set and constructs a hint–accuracy collection $C_x$ for each problem $x$.  
Then, SEELE performs $m$ rounds of generation.  
In each round, $C_x$ is augmented by adding the margin points $(0,0)$ and $(1,1)$ whenever $p_x=0$ or $p_x=1$ has not yet been evaluated.  
This augmentation substantially enhances fitting accuracy and stability during the early rounds.  
Subsequently, the predictor $f_\phi$ is optimized on the augmented $\hat C_x$ and used to estimate the optimal hint length $l$.  
With the estimated hint length, SEELE invokes the policy model to generate $n$ rollouts for the hinted problem $\hat{x}=x\oplus y_{1:l}$ and computes the corresponding rewards.  
These rewards are then aggregated to calculate the intra-round accuracy $a_x$, which is used to update $C_x$.  
Finally, after $m$ rounds of rollouts, SEELE computes the advantages over all $mn$ outputs following the Dr.GRPO formulation~\citep{liuUnderstandingR1ZeroLikeTraining2025}, and updates the policy model according to Eq.~\eqref{eq:impl-loss}.

\subsection{Computation Overhead of SEELE}\label{apsec:overhead}
Although adopting a multi-turn rollout strategy, SEELE has nearly identical FLOPs to other RL methods (under the same number of rollouts). The only additional cost over GRPO comes from multi-round sampling because the prompt KV-cache is not shared across rounds. The extra cost from the non-linear least squares fitting is negligible, as the computation is lightweight (3 parameters) and can be parallelized across rollouts on CPU cores.

To provide a quantitative comparison, we report the FLOPs of each phase in a single RL step in Table~\ref{tab:flops}. Following the prior work~\citep{kaplan_scaling_2020}, we approximate the FLOPs of the forward and backward passes as $2NL$ and $4NL$, respectively, where $N$ denotes the amount of LLM's non-embedding parameters and $L$ denotes the sequence length.

\setlength\tabcolsep{10pt}
\begin{table}[h]
    \footnotesize
    \centering
    \caption{FLOPs comparison with GRPO.}\label{tab:flops}
    \begin{tabular}{lcc}
\toprule
 Phase & GRPO & SEELE \\
\midrule
Rollout 
& $2NL_p + 2mn N L_r$
& $2mN L_p + 2mn N L_r$ \\
Old Prob. 
& $2mnN(L_p+L_r)$
& $2mnN(L_p+L_r)$ \\
Ref. Prob. 
& $2mnN(L_p+L_r)$
& $2mnN(L_p+L_r)$ \\
Update 
& $6mnN(L_p+L_r)$
& $6mnN(L_p+L_r)$ \\
\midrule
Total 
& \makecell{$2NL_p + 2mnN L_r$ \\ $+\,10mnN(L_p+L_r)$}
& \makecell{$2mN L_p + 2mnN L_r$ \\ $+\,10mnN(L_p+L_r)$} \\
\bottomrule
\end{tabular}
\end{table}

Under our experimental configuration, $m=4,n=8,L_p=1024,L_r=2048$, we obtain:
$$\frac{FLOPs(\text{SEELE})}{FLOPs(\text{GRPO})}=1.0055.$$
Thus, the FLOPs increment introduced by multi-round sampling is only about $0.55\%$, which is effectively negligible.

\subsection{Training Data Synthesis}\label{apsec:data-synth}
Our data synthesis procedure consists of two phases: filtering and annotation.  
We use DeepMath-103K~\citep{heDeepMath103KLargeScaleChallenging2025} as the initial dataset.  
First, to construct a more challenging subset, we employ Qwen2.5-7B~\citep{qwenQwen25TechnicalReport2025} to sample 8 reasoning traces per instance (temperature $=0.6$, maximum length $=2048$) and retain only those instances for which all traces are incorrect.  
Second, we use DeepSeek-V3~\citep{deepseek-aiDeepSeekV3TechnicalReport2025} to generate step-by-step reasoning annotations.  
The annotation prompt is presented as follows.  
We instruct DeepSeek-V3 to produce a logically complete and concise solution based on the reference solution provided in the original dataset.

\begin{mybox}{\textbf{Step-by-Step Annotation Prompt}}

Task: Generate a clear, step-by-step, and complete solution to the following problem with these requirements:

1. \textbf{Five or fewer Numbered Steps:}

\quad Ensure no logical jumps—every non-trivial inference must be justified.

2. \textbf{Key Explanations Included:}

\quad a. Briefly explain ``why'' for non-obvious steps (e.g., ``We use X method because...'').

\quad b. Avoid redefining terms/concepts already introduced.

3. \textbf{Full Calculations:}

\quad a. Show at least one intermediate step for computations (e.g., a = b + c → a = 5 + 3 = 8).

\quad b. For symbolic math, state the rules/theorems used (e.g., ``By the chain rule...'').

4. \textbf{Final Answer:}

\quad Mark clearly with \textbackslash boxed\{\}.

\textbf{Original Question:}\\
\{\textit{question}\}

\textbf{Reference Solution (for guidance only):}\\
\{\textit{reference\_solution}\}

Begin your new solution:
\end{mybox}

\subsection{Chain-of-Thought Prompt Template}\label{apsec:cot-prompt}
Following previous work~\citep{liuUFTUnifyingSupervised2025}, we adopt the DeepSeek-R1 prompt template for both training and evaluation.  
This template contains minimal special tokens and is well-suited for the base model.  
When generation is complete, we first attempt to extract the text enclosed within \verb|<answer> ... </answer>|, which is then used for subsequent answer extraction.  
If no match is found, we instead extract the content within the \textbackslash boxed\{\} tag.

\begin{mybox}{\textbf{Chain-of-Thought Prompt}}

\begin{verbatim}
    
\end{verbatim}
A conversation between User and Assistant. The user asks a question, and the Assistant solves it. The assistant first thinks about the reasoning process in the mind and then provides the user with the answer.

User: \{\textit{question}\}

Show your work in \verb|<think> </think>| tags. And return the final answer in \verb|<answer> </answer>| tags, for example \verb|<answer>| 12 \verb|</answer>|.

Assistant: Let me solve this step by step.

\verb|<think>|

\{\textit{completion}\}

\end{mybox}

\section{Extended Empirical Analysis}
\subsection{Performance on Various Models}\label{sec:more-model}
In addition to Qwen2.5-Math-7B and Qwen2.5-1.5B, we further evaluate SEELE on three base models from different families: Qwen2.5-3B, LLaMA~3.2-3B, and Mathstral-7B-v0.1, and summarize the results in Table~\ref{tab:supp}.
Overall, SEELE consistently delivers the strongest performance across both the math reasoning and general domain reasoning benchmarks, demonstrating its wide applicability.

\setlength\tabcolsep{2pt}
\begin{table*}[h]
    \centering
    \caption{Performance on math and general domain reasoning benchmarks using Qwen2.5-3B, LLaMA 3.2-3B, and Mathstral-7B-v0.1.}\label{tab:supp}
    \scriptsize
    \begin{tabular}{lccccccc|cccc}
        \toprule
        \multirow{2}{*}{\textbf{Model}} & \multicolumn{7}{c}{\textbf{Math Reasoning}} & \multicolumn{4}{c}{\textbf{General Domain Reasoning}}\\
        \cmidrule{2-8}\cmidrule{9-12}
        & \textbf{GSM8K} & \textbf{MATH500} & \textbf{Minerva} & \textbf{Olympiad} & \textbf{AIME24} & \textbf{AMC23} & \textbf{Avg.} & \textbf{ARC-C} & \textbf{GPQA-D} & \textbf{MMLU-Pro} & \textbf{Avg.}\\
        \midrule
        \textbf{Qwen2.5-3B} & 73.3 & 29.0 & 7.0 & 10.7 & 0.6 & 11.6 & 22.0 & 66.6 & 23.7 & 15.2 & 35.2\\
        + SFT & 73.1 & 52.6 & 18.4 & 18.5 & 2.3 & 25.0 & 31.7 & 75.7 & 30.3 & 40.1 & 48.7\\
        + GRPO & 78.5 & 45.8 & 17.3 & 15.0 & 1.7 & 23.9 & 30.4 & 75.6 & \textbf{36.9} & 40.5 & 51.0\\
        + LUFFY & 81.1 & 56.6 & 21.0 & 21.3 & 2.7 & 27.0 & 35.0 & 78.4 & 32.8 & 41.5 & 50.9\\
        + UFT & 84.8 & 62.6 & 22.1 & 25.9 & 5.8 & \textbf{41.6} & 40.5 & 79.4 & 35.4 & 42.9 & 52.6\\
        + Prefix-RFT& 77.4 & 57.0 & 21.3 & 21.9 & \textbf{6.0} & 31.5 & 35.9 & \textbf{82.2} & 35.9 & 37.9& 52.0\\
        \rowcolor{gray!10}+ SEELE (Ours)& \textbf{86.3} & \textbf{66.4} & \textbf{26.1} & \textbf{28.9} & 5.9 & 39.4 & \textbf{42.2} & 81.2 & 34.3 & \textbf{44.0} & \textbf{53.2}\\
        \midrule
        \textbf{LLaMA 3.2-3B} & 8.1 & 4.2 & 2.6 & 1.9 & 0.0 & 2.7 & 3.3 & 13.7 & 15.2 & 7.3 & 12.1\\
        + SFT & 20.7 & 8.8 & 4.0 & 3.0 & 0.2 & 4.0 & 6.8 & 59.0 & 27.3 & \textbf{26.3} & 37.5\\
        + GRPO & 5.9 & 8.2 & 5.5 & 2.5 & 0.0 & 3.2 & 4.2 & 43.5 & 28.8 & 23.9 & 32.1\\
        + LUFFY & 11.9 & 9.6 & 4.8 & 2.4 & 0.2 & 3.5 & 5.4 & 58.8 & \textbf{30.8} & 26.0 & 38.5\\
        + UFT & 24.2 & 11.4 & \textbf{8.1} & 3.6 & 0.2 & \textbf{7.5} & 9.2 & 56.2 & 27.3 & 25.2 & 36.2\\
        + Prefix-RFT& 21.1 & 9.2 & 5.8 & 3.6 & 0.1 & 2.8 & 7.1 & 55.5 & 28.4 & 24.9 & 36.3\\
        \rowcolor{gray!10}+ SEELE (Ours)& \textbf{29.5} & \textbf{12.8} & 7.4 & \textbf{5.2} & \textbf{0.7} & 6.5 & \textbf{10.4} & \textbf{62.4} & 28.8 & 26.1 & \textbf{39.1}\\
        \midrule
        \textbf{Mathstral-7B} & 76.0 & 34.6 & 15.8 & 14.5 & 1.4 & 15.7 & 26.3 & 62.9 & 29.3 & 16.7 & 36.3\\
        + SFT & 80.1 & 53.4 & 24.6 & 21.6 & 1.8 & 26.6 & 34.7 & 73.1 & \textbf{44.4} & 42.2 & 53.2\\
        + GRPO & 85.6 & 44.0 & 22.4 & 16.0 & 0.9 & 21.7 & 31.8 & 80.0 & 38.4 & 43.5 & 54.0\\
        + LUFFY & 88.4 & 60.0 & 25.0 & 24.4 & 6.4 & 32.4 & 39.4 & 80.2 & 32.8 & 44.1 & 52.4\\
        + UFT & 87.9 & 57.6 & 23.2 & 20.6 & 4.3 & 33.1 & 37.8 & 81.8 & 37.9 & 47.7 & 55.8\\
        + Prefix-RFT& 86.3 & 59.6 & 24.3 & 22.5 & 3.6 & 31.7 & 38.0 & 79.7 & 35.4 & 46.2 & 53.8\\
        \rowcolor{gray!10}+ SEELE (Ours)& \textbf{90.2} & \textbf{63.4} & \textbf{27.9} & \textbf{29.5} & \textbf{6.7} & \textbf{38.4} & \textbf{42.7} & \textbf{82.9} & 38.9 & \textbf{50.5} & \textbf{57.4}\\
        \bottomrule
    \end{tabular}
\end{table*}

\textbf{Qwen2.5-3B.}\quad
In the in-domain mathematical reasoning setting, SEELE achieves an average improvement of \textbf{+11.8} points over GRPO and \textbf{+10.5} points over SFT, surpassing the best baseline by 1.7\%.
The improvement is consistent across all six math benchmarks, underscoring the benefits of introducing off-policy demonstrations and dynamically adapting data difficulty.
We further observe that GRPO generally underperforms SFT on complex reasoning tasks (e.g., MATH500, AIME24), while showing comparable or slightly better performance on relatively easy tasks (e.g., GSM8K, ARC-C), highlighting the limitations of purely on-policy exploration.
Moreover, the relatively low performance of GRPO and SFT indicates that exclusive self-exploration or pure imitation alone is insufficient to cultivate strong complex reasoning capabilities.
In the out-of-domain general reasoning setting, SEELE demonstrates strong generalization, achieving an average improvement of \textbf{+2.2} points over GRPO, while other baselines exhibit only comparable performance.

\textbf{LLaMA~3.2-3B.}\quad
A similar trend is observed on the smaller LLaMA~3.2-3B model, where overall math reasoning is more challenging.
Despite the low base performance, SEELE improves the math average to \textbf{10.4}, outperforming SFT and GRPO by \textbf{+3.6} and \textbf{+6.2} points respectively.
Notably, GRPO again fails to consistently surpass SFT, confirming that naive on-policy reinforcement struggles with scarce reasoning signals.
Other supervision-aided RL methods (LUFFY, UFT, Prefix-RFT) provide moderate gains, showing incompetence on the weak initial model.
For general domain reasoning, SEELE also reaches the highest average (39.1), exceeding GRPO by \textbf{+7.0} and SFT by \textbf{+1.6}, showing that our strategy scales effectively even for smaller and weaker base models.

\textbf{Mathstral-7B.}\quad
On the larger Mathstral-7B, which already exhibits strong math ability, SEELE continues to be the state-of-the-art.
It raises the math average to \textbf{42.7\%}, surpassing SFT by \textbf{+8.0} and GRPO by \textbf{+10.9} points, and outperforming the strongest competing RL-with-supervision baseline (LUFFY) by \textbf{+3.3} points.
In challenging tasks such as Olympiad and AIME24, SEELE achieves the largest absolute gains, demonstrating its ability to handle complex reasoning even when the base model is powerful.
For general domain reasoning, SEELE further improves the average to \textbf{57.4\%}, a \textbf{+1.6\%} gain over the best baseline, again highlighting superior out-of-domain generalization.

In summary, SEELE is generally effective for training models of different families (Qwen, LLaMA, and Mathstral) and sizes (1.5B, 3B, and 7B), showing great robustness.
Notably, for models with very weak math ability (e.g., LLaMA 3.2-3B), SEELE still achieves considerable improvement.

\subsection{Performance on Strong Reasoning and Instruction Models}\label{sec:strong}
\setlength\tabcolsep{1pt}
\begin{table*}[t]
    \centering
    \caption{Performance on math and general domain reasoning benchmarks using DeepSeek-R1-Distill-Qwen-7B and Qwen3-4B-Instruct-2507.}
    \label{tab:strong}
    \scriptsize
    \begin{tabular}{lccccccc|cccc}
        \toprule
        \multirow{2}{*}{\textbf{Model}} & \multicolumn{7}{c}{\textbf{Math Reasoning}} & \multicolumn{4}{c}{\textbf{General Domain Reasoning}}\\
        \cmidrule{2-8}\cmidrule{9-12}
        & \textbf{GSM8K} & \textbf{MATH500} & \textbf{Minerva} & \textbf{Olympiad} & \textbf{AIME24} & \textbf{AMC23} & \textbf{Avg.} & \textbf{ARC-C} & \textbf{GPQA-D} & \textbf{MMLU-Pro} & \textbf{Avg.}\\
        \midrule

        \textbf{DeepSeek-R1-Distill-Qwen-7B} & 89.5 & 72.6 & 25.7 & 37.9 & 17.3 & 53.3 & 49.4 & 69.7 & 34.8 & 30.7 & 45.1\\
        + GRPO & 90.1 & 84.4 & 31.2 & 51.1 & 27.9 & 75.5 & 60.0 & 75.9 & \textbf{44.4} & 51.1 & 57.1\\
        \rowcolor{gray!10}
        + SEELE (Ours) & \textbf{90.3} & \textbf{87.2} & \textbf{34.9} & \textbf{53.0} & \textbf{38.9} & \textbf{79.0} & \textbf{63.9} & \textbf{79.1} & 42.4 & \textbf{53.6} & \textbf{58.4}\\

        \midrule

        \textbf{Qwen3-4B-Instruct-2507} & 94.1 & 86.6 & 35.7 & 49.0 & 29.4 & 75.2 & 61.7 & 90.2 & 52.0 & 60.0 & 67.4\\
        + SFT & 83.5 & 65.2 & 30.1 & 28.3 & 7.2 & 41.0 & 42.6 & 89.5 & 45.5 & 60.2 & 65.1\\
        + GRPO & 93.7 & 90.4 & \textbf{39.7} & 55.4 & 38.6 & 87.3 & 67.5 & \textbf{92.7} & 55.0 & 70.5 & 72.7\\
        \rowcolor{gray!10}
        + SEELE (Ours) & \textbf{94.5} & \textbf{92.0} & \textbf{39.7} & \textbf{56.1} & \textbf{40.9} & \textbf{88.0} & \textbf{68.5} & 92.6 & \textbf{58.1} & \textbf{70.8} & \textbf{73.8}\\

        \bottomrule
    \end{tabular}
\end{table*}
Table~\ref{tab:strong} further demonstrates that SEELE remains effective beyond base models, consistently improving already strong reasoning and instruction-tuned models. On DeepSeek-R1-Distill-Qwen-7B, a capable reasoning model, SEELE surpasses GRPO by a clear margin, improving Math Avg. from 60.0\% to 63.9\% and General Avg. from 57.1\% to 58.4\%, with especially notable gains on more challenging benchmarks such as AIME24 (+11.0\%) and Minerva (+3.7\%). This suggests that SEELE can further unlock latent reasoning potential even when the model already possesses strong chain-of-thought capabilities. 

Similarly, on Qwen3-4B-Instruct-2507, a high-quality instruction-tuned model, SEELE consistently outperforms both SFT and GRPO baselines, achieving the best overall Math Avg. (68.5\%) and General Avg. (73.8\%). The gains are particularly evident on difficult mathematical reasoning tasks such as MATH500, Olympiad, and AIME24. These results indicate that SEELE is not limited to weak or untuned base models, but can also serve as an effective zero-RL enhancement for advanced instruction-following models. Overall, the strong performance across these diverse starting points highlights the robustness and broad applicability of SEELE.

\subsection{Pass@k Evaluation and Diversity Analysis}\label{sec:passk}

To further examine whether SEELE merely follows a single hinted trajectory or truly improves the underlying reasoning policy, we evaluate all methods using the Pass@k metric with $k \in \{1,4,8,16\}$. Pass@k measures whether at least one correct solution appears among $k$ independently sampled rollouts, and thus reflects both single-sample accuracy and the diversity of successful reasoning paths. If gains only arise from collapsing onto one narrow solution mode induced by hints, improvements would mainly appear in Pass@1 while higher-$k$ gains would be limited. In contrast, consistent improvements across all $k$ values indicate that the model discovers a broader and more effective set of reasoning strategies.

\begin{table}[t]
\centering
\footnotesize
\setlength{\tabcolsep}{10pt}
\caption{Pass@k comparison across math and general reasoning benchmarks using Qwen2.5-Math-7B. SEELE consistently achieves the best performance for all $k$ values.}
\label{tab:passk}
\begin{tabular}{lcccccccc}
\toprule
\multirow{2}{*}{\textbf{Method}} & \multicolumn{4}{c}{Math Average} & \multicolumn{4}{c}{General Average} \\
\cmidrule(lr){2-5} \cmidrule(lr){6-9}
& P@1 & P@4 & P@8 & P@16 & P@1 & P@4 & P@8 & P@16 \\
\midrule
GRPO         & 55.7 & 61.8 & 64.2 & 66.3 & 53.2 & 57.2 & 58.6 & 63.1 \\
UFT          & 55.6 & 60.7 & 63.2 & 65.3 & 57.3 & 64.1 & 66.0 & 67.1 \\
Prefix-RFT   & 56.3 & 62.1 & 64.8 & 67.5 & 55.8 & 60.1 & 61.4 & 62.2 \\
SEELE (Ours) & \textbf{59.0} & \textbf{64.5} & \textbf{66.0} & \textbf{69.6} & \textbf{58.5} & \textbf{64.9} & \textbf{67.0} & \textbf{68.2} \\
\bottomrule
\end{tabular}
\end{table}

As shown in Table~\ref{tab:passk}, SEELE consistently outperforms all baselines across both domains and all sampling budgets. On math tasks, SEELE improves over GRPO from $55.7\%$ to $59.0\%$ at Pass@1, and maintains an advantage at larger sampling budgets, reaching $69.6\%$ at Pass@16. This suggests that SEELE not only increases single-rollout correctness, but also enhances the probability of generating multiple valid solution trajectories.

A similar trend is observed on general reasoning tasks. SEELE attains the strongest Pass@1 performance ($58.5\%$), while maintaining leading results at Pass@4, Pass@8, and Pass@16. Notably, methods that rely on stronger supervised prefixes or direct fine-tuning may improve some low-$k$ settings, but their gains are less consistent as $k$ increases, implying more limited exploration diversity.

Overall, these results support our central hypothesis: strategic hints do not simply steer the model toward one predefined reasoning path. Instead, they reshape the optimization landscape so that reinforcement learning can discover a wider set of high-quality reasoning behaviors.

\section{Rollout Manipulation Analysis}

\subsection{Target Accuracy Analysis}
\begin{figure*}[t]
    \centering
    \includegraphics[width=0.95\linewidth]{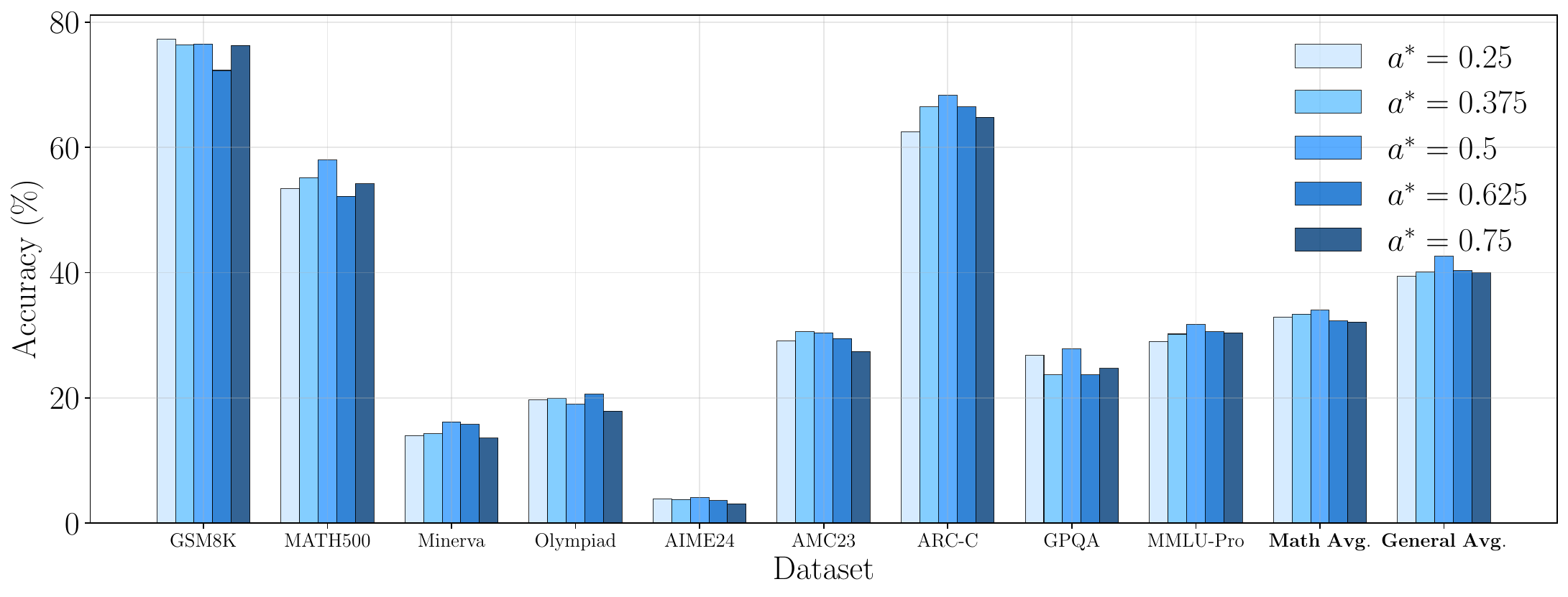}
    \caption{Performance for different target accuracies $a^*$ using Qwen2.5-1.5B.}
    \label{fig:target_acc}
\end{figure*}

To verify the critical role of rollout accuracy in RLVR, we conduct an ablation study by varying the target accuracy $a^*$ and examining its impact on performance. Specifically, we set $a^*$ to values in the range $\{i/n \mid i=2,3,\ldots,n-2\}$, while excluding the boundary cases $i=1$ and $i=n-1$ to avoid situations where rollouts become entirely incorrect or entirely correct.
The results, summarized in Figure~\ref{fig:target_acc}, lead to the following observations:  
(1) Setting the target accuracy to $0.5$ yields the best performance, while performance degrades gradually as $a^*$ deviates from this value;  
(2) The degradation trend is approximately symmetric, whether the target accuracy is increased or decreased.  
These findings are consistent with both intuition and theoretical analysis: data that is either overly difficult or overly simple hinders effective training. For completeness, we also report detailed accuracy control results in Appendix~\ref{sec:target_control}, which further demonstrate that our method can precisely regulate the rollout accuracy to any desired target value.

\begin{figure}[t]
    \centering
    \includegraphics[width=0.7\linewidth]{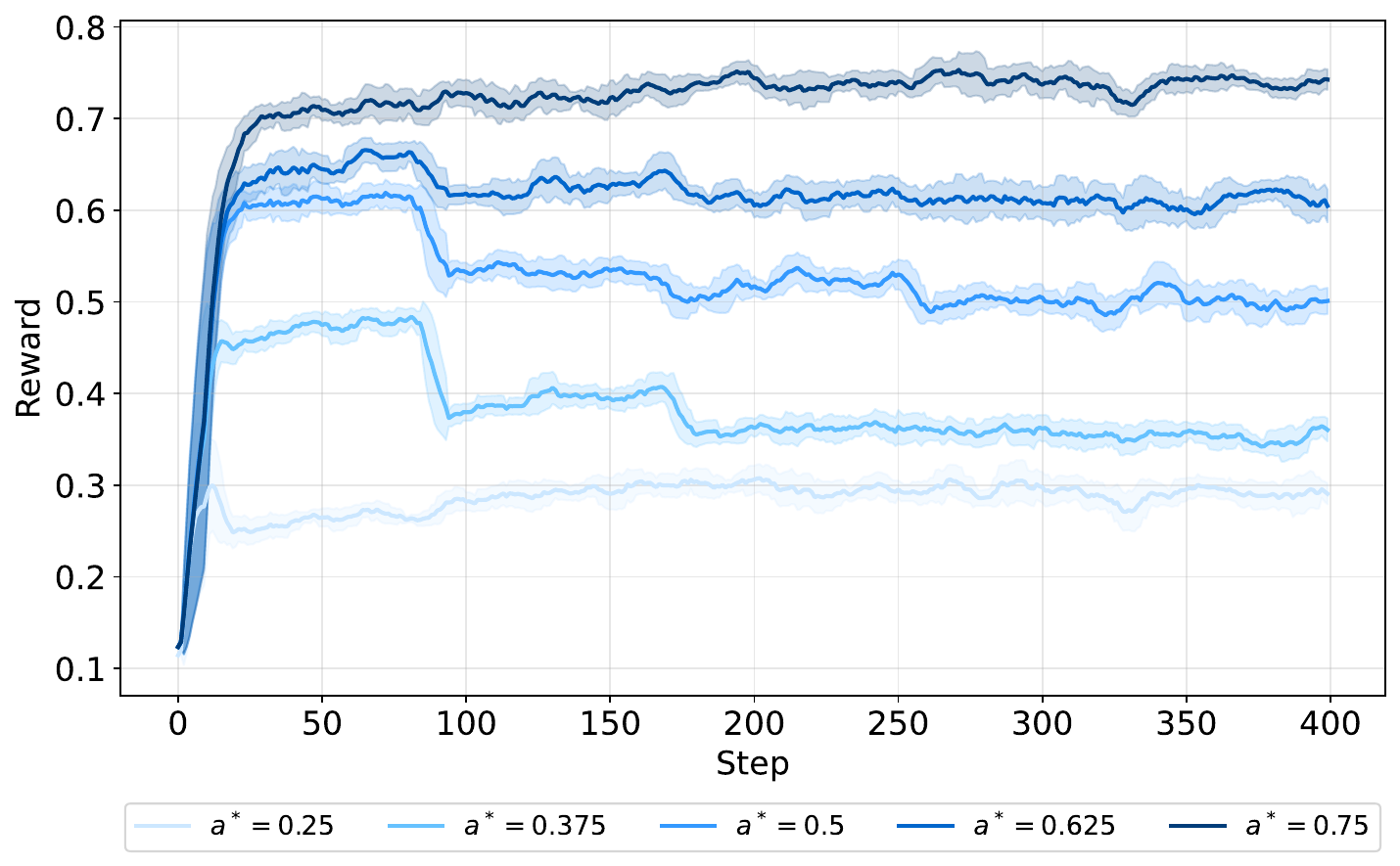}
    \caption{Reward across the training steps for various target accuracies.}
    \label{fig:target_control}
\end{figure}

\subsection{Rollout Accuracy Manipulation}\label{sec:target_control}
To verify that our multi-round sampling framework can arbitrarily manipulate the rollout accuracy, we set the target accuracy to $i/n,i=2,3,\cdots,n-2$.
We exclude $1/n$ and $(n-1)/n$ because they are too close to the boundary, which easily leads to all wrong/correct rollout sampling.
We set $n=8$ following the main experiment setup.
The results are shown in Figure~\ref{fig:target_control}.
After the cold start for an epoch, the reward converges to the target accuracy across all settings with very little error (less than 0.02 for $a^*>0.25$), showing a stable trend until the end of the training.
The fluctuation and deviation diminish as the training progresses.
The results of $a^*=0.25$ are slightly higher than the target because they touch the difficulty lower bound, where the policy model can generate sufficiently correct outputs without the aid of hints.

\setlength\tabcolsep{2pt}
\begin{table*}[t]
    \centering
    \caption{Performance of different multi-round configurations using Qwen2.5-1.5B.}
    \label{tab:round_config}
    \scriptsize
    \begin{tabular}{ccccccccc|cccc}
        \toprule
        \multicolumn{2}{c}{\textbf{Multi-Round}} & \multicolumn{7}{c}{\textbf{Math Reasoning}} & \multicolumn{4}{c}{\textbf{General Domain Reasoning}}\\
        \cmidrule{3-9}\cmidrule{10-13}
        \multicolumn{2}{c}{\textbf{Rollout Scheme}}& \textbf{GSM8K} & \textbf{MATH500} & \textbf{Minerva} & \textbf{Olympiad} & \textbf{AIME24} & \textbf{AMC23} & \textbf{Avg.} & \textbf{ARC-C} & \textbf{GPQA-D} & \textbf{MMLU-Pro} & \textbf{Avg.}\\
        \midrule
        $mn=8$ & $m=2$ & 76.0& 50.8& 16.2& 18.2& 3.0& 28.1& {32.1}& 65.2& {26.3}& 28.1& {39.9}\\
        \midrule
        $mn=16$ & $m=4$ & {76.8}& 51.0& 15.4& 19.0& 3.3& 28.4& {32.3}& 66.3& {24.7}& 29.7& {40.2}\\
        \midrule
        \multirow{3}{*}{$mn=24$} & $m=3$ & 75.1& 53.4& 14.7& \textbf{20.0}& \textbf{4.6}& 31.8& 33.3& 67.9& 21.9& 29.8& 39.9\\
        & $m=4$ & 77.1& 54.4& \textbf{16.9}& 19.9& 3.0& \textbf{32.1}& 33.9& 66.6& {24.7}& \textbf{33.2}& 41.5\\
        & $m=6$ & \textbf{77.6}& 53.8& 15.8& 18.5& 2.9& 30.5& {33.2}& 68.9& 23.8& 33.0& {41.9}\\
        \midrule
        \multirow{2}{*}{$mn=32$} & $m=4$ & {76.5}& \textbf{58.0}& 16.2& 19.9& 4.1& 30.4& \textbf{34.2}& 68.3& \textbf{27.8}& 31.7& {\textbf{42.6}}\\
        & $m=8$ & 76.9& 54.2& {16.5}& 19.3& 3.3& 31.4& 33.6& \textbf{69.8}& 23.2& 32.4& 41.8\\
        \bottomrule
    \end{tabular}
\end{table*}

\subsection{Rollout Scheme Analysis}\label{sec:scheme}
Given a fixed total number of rollouts, the number of rounds can be configured in multiple ways.
Increasing the number of rounds provides more samples for regression, but it reduces the number of samples per round, which in turn increases the variance in estimating accuracy for a specific hinting rate.
To examine the actual effect of the multi-round rollout configuration and identify the optimal setting, we train Qwen2.5-1.5B with varying numbers of rollout rounds under different total rollout budgets.
The results are presented in Table~\ref{tab:round_config}.
Our observations are as follows:  
(1) The performance of SEELE is relatively insensitive to the specific rollout configuration;  
(2) A three-round scheme underperforms the four-round scheme on both domain averages, as only two sample points are available for constructing the three-parameter logistic (3PL) model, which is insufficient for accurate estimation;  
(3) A four-round scheme generally achieves the highest performance. Increasing the number of rounds beyond four reduces the number of samples per round, leading to higher estimation variance in single-round accuracy.
Based on these findings, we therefore select $m = 4$ as a practical choice for optimal performance.

\begin{figure*}[t]
    \centering
    \includegraphics[width=0.84\linewidth]{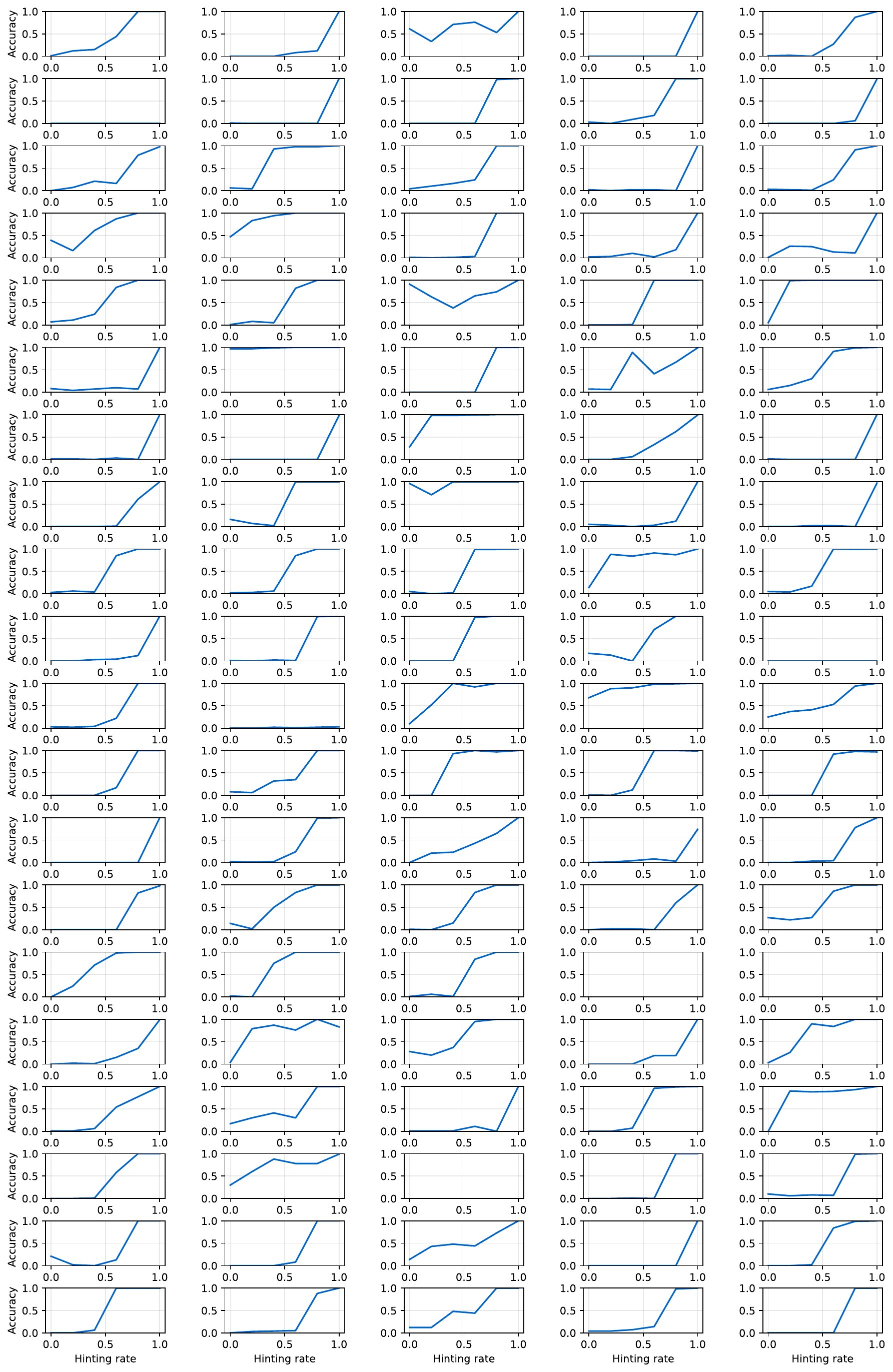}
    \caption{Accuracy with respect to the hinting rate for 100 training examples using Qwen2.5-3B.}
    \label{fig:acc_hint_full}
\end{figure*}

\subsection{Relationship between Prediction Accuracy and Hinting Rate}\label{sec:acc_hint_full}
We analyze the relationship between prediction accuracy and hinting rate using the Qwen2.5-3B checkpoint after 30 training steps, corresponding to the early rising stage of learning.
For each example, we enumerate the hint length and prompt the model to complete the hinted problem.
Accuracy is computed over 100 randomly sampled traces with a temperature setting of 1.

Figure~\ref{fig:acc_hint_full} illustrates the accuracy–hinting rate curves for these 100 examples.
In most cases, the curves exhibit an S-shaped trend: accuracy remains close to zero until a critical proportion of the solution is revealed, after which it rises rapidly to nearly 1.
This pattern is highly consistent with the predictions of the three-parameter logistic (3PL) model.

To quantitatively evaluate the fitness of different hint-length modeling strategies, we calculate the coefficient of determination ($R^2$) and root mean squared error (RMSE) over the training set using Qwen2.5-Math-7B as the evaluator model. For each hint length, we generate 100 rollouts to approximate the underlying performance curve, and then fit different predictive models to the observed results.

\begin{wraptable}{r}{0.42\textwidth}
\centering
\setlength{\tabcolsep}{10pt}
\caption{Goodness-of-fit comparison.}
\label{tab:fit_result}
\begin{tabular}{lcc}
\toprule
Method & $R^2$ & RMSE \\
\midrule
Linear & 0.694 & 0.164 \\
3PL    & \textbf{0.929} & \textbf{0.031} \\
\bottomrule
\end{tabular}
\end{wraptable}

As shown in Table~\ref{tab:fit_result}, the 3PL model achieves the strongest goodness-of-fit, with an $R^2$ of $0.929$ and an RMSE of only $0.031$, significantly surpassing the Linear alternative.
These results indicate that the true relationship between hint proportion and model performance is highly nonlinear, and cannot be adequately captured by a simple linear approximation.
Overall, the superior fit quality of 3PL provides strong empirical support for SEELE's dynamic adjustment mechanism. By more accurately estimating the capability-aware optimal hint ratio, SEELE can better match problem difficulty with model competence, leading to improved training efficiency and downstream reasoning performance.


\end{document}